\pdfoutput=1

\documentclass[11pt]{article}

\usepackage[final]{acl}

\usepackage{times}
\usepackage{latexsym}

\usepackage[T1]{fontenc}

\usepackage[utf8]{inputenc}

\usepackage{microtype}

\usepackage{inconsolata}

\usepackage{graphicx}
\usepackage{subcaption}
\usepackage{booktabs}
\usepackage{amsmath}
\usepackage{graphbox}
\usepackage{bbm}
\usepackage{amssymb}
\usepackage{makecell}

\usepackage{titlesec}
\titlespacing*{\paragraph}{0ex}{0.5ex}{1ex}

\title{Demystifying Verbatim Memorization in Large Language Models}

\author{Jing Huang, Diyi Yang$\textsuperscript{\textasteriskcentered}$, Christopher Potts\thanks{Equal advising.}  \\
 Stanford University \\
  \texttt{ \{hij, diyiy, cgpotts\}@stanford.edu} }

\definecolor{x11purple}{RGB}{160, 32, 240}
\definecolor{forestgreen}{RGB}{34, 139, 34}
\definecolor{brickred}{RGB}{203, 65, 84}
\definecolor{brass}{RGB}{225, 193, 110}

\newcommand{\seq}{\textbf{x}}
\newcommand{\basemodel}{\mathcal{M}}
\newcommand{\controlmodel}{\mathcal{M}^{(\varnothing)}}
\newcommand{\treatmodel}{\mathcal{M}^{(X)}}
\newcommand{\pythia}[1]{\texttt{pythia-{#1}-deduped}}
\newcommand{\E}[2]{${#1}\times 10^{#2}$}

\begin{document}

\maketitle
\begin{abstract}

Large Language Models (LLMs) frequently memorize long sequences verbatim, often with serious legal and privacy implications. Much prior work has studied such verbatim memorization using observational data. To complement such work, we develop a framework to study  verbatim memorization in a controlled setting by continuing pre-training from Pythia checkpoints with injected sequences. We find that (1) non-trivial amounts of repetition are necessary for verbatim memorization to happen; (2) later (and presumably better) checkpoints are more likely to verbatim memorize sequences, even for out-of-distribution sequences; (3) the generation of memorized sequences is triggered by distributed model states that encode high-level features and makes important use of general language modeling capabilities. Guided by these insights, we develop stress tests to evaluate unlearning methods and find they often fail to remove the verbatim memorized information, while also degrading the LM. Overall, these findings challenge the hypothesis that verbatim memorization stems from specific model weights or mechanisms. Rather, verbatim memorization is intertwined with the LM's general capabilities and thus will be very difficult to isolate and suppress without degrading model quality.

\end{abstract}

\section{Introduction}

Verbatim memorization refers to LLMs outputting long sequences of texts that are exact matches of training examples \cite{carlini2021extracting,carlini2023quantifying}. Unlike recalling factual knowledge or fixed expressions, verbatim memorization can have serious copyright and privacy implications \cite{karamolegkou-etal-2023-copyright,chen2024copybench,lee2023plagiarize,carlini2021extracting,shokri2017membership} and potentially waste model capacity~\cite{nasr2023scalable}. Recent work has identified data frequency and model size as factors contributing to verbatim memorization in LLMs~\cite{carlini2023quantifying,prashanth2024recitereconstruct,karamolegkou-etal-2023-copyright}. However, it is still not well understood why and how LLMs verbatim memorize certain texts in training data.

\begin{figure}[t!]

    \centering
    \includegraphics[width=0.85\linewidth,trim={0 3cm 9cm 0},clip]{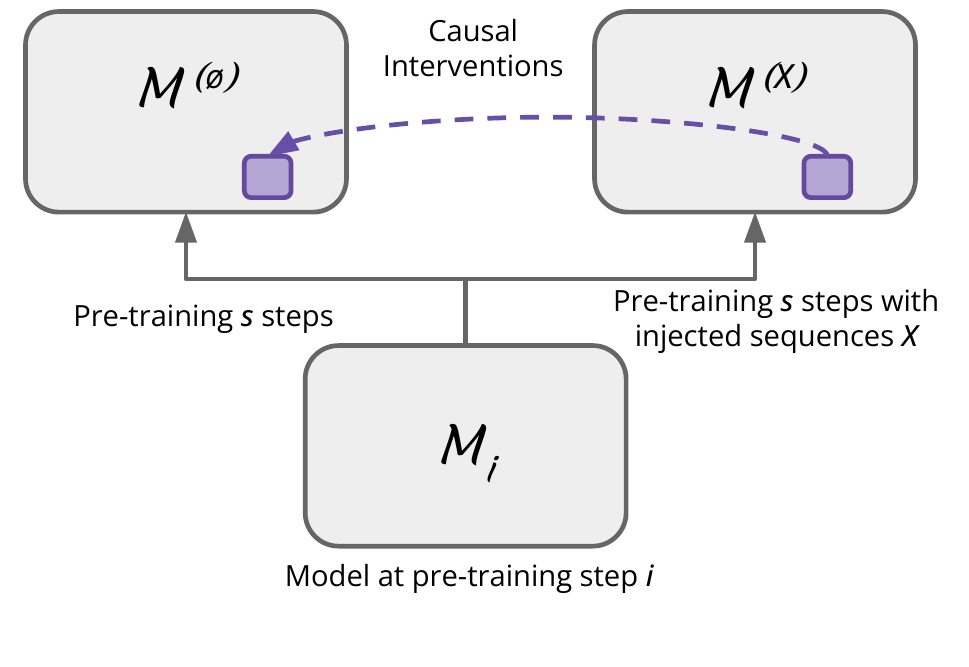}
    \vspace{-4ex}
    \caption{An overview of our sequence injection framework, which creates a control model $\controlmodel$ and a treatment model $\treatmodel$ by continued pre-training from the same checkpoint, with a set of sequences to memorize $X$ injected into $\treatmodel$'s training data. Our framework explicitly creates a counterfactual state that allows us to study, via causal interventions, what the model would have been if it had not seen a particular sequence.}
    \vspace{-4ex}
    \label{fig:framework}
\end{figure}

One hypothesis is that \emph{there are specialized model weights or mechanisms dedicated to recalling the verbatim memorized texts}~\cite{nasr2023scalable,chang2024localization,stoehr2024localizing}. Under this view, preventing verbatim memorization should be straightforward. For example, localizing and intervening on these dedicated components (e.g., a few neurons;~\citealt{chang2024localization,maini2023can} or a particular attention head; \citealt{stoehr2024localizing}) should remove verbatim memorized texts while preserving model quality.
However, recent work indicates that removing verbatim memorized information is challenging. Preventing verbatim memorization during decoding does not stop variants  of the memorized texts from being generated \cite{ippolito-etal-2023-preventing}, and memorized texts can be retrieved in contexts different from the ones seen in training~\cite{karamolegkou-etal-2023-copyright,ippolito-etal-2023-preventing,kassem2024alpaca}. In addition, unlearning via fine-tuning and pruning degrades model quality \cite{stoehr2024localizing,chang2024localization,maini2024tofu,lynch2024eightmethod,chen2023unlearn}. These findings suggest an alternative view: rather than having specialized weights or mechanisms dedicated to verbatim memorization, \emph{models might be reconstructing memorized sequences using  features learned from general language modeling}. This would explain why we are unable to localize memorized texts and why mitigating memorization can fundamentally alter model behaviors.

In this paper, we seek to answer these questions. We develop a framework for studying memorization in controlled settings: given an LM checkpoint $\basemodel$, we continue pre-training $\basemodel$ on the original training data but with specific novel sequences inserted at controlled frequencies. 
This framework complements existing observational methods, and allows us to decouple factors that potentially affect memorization, including model size, data frequency, and model quality. We use this framework in experiments with the Pythia family of models~\citep{biderman2023pythia}. Our core findings are as follows: (1) Sequences need to be repeated a non-trivial number of times to be memorized. The perception that a model verbatim memorizes a sequence that occurs once in pre-training is likely an illusion. (2) Later (and presumably better) checkpoints are more likely to verbatim memorize sequences, and even out-of-domain sequences are memorized at non-trivial rates by the best models. (3) Only some tokens in verbatim memorized sequences causally depend on a set of distributed triggering states that encode high-level semantic features, with the rest produced by regular LM decoding. Based on these findings, we develop stress tests to evaluate unlearning methods and find they often fail to remove verbatim memorized information while also degrading model quality.

Overall, these results challenge the view that verbatim memorization stems from specific model weights or mechanisms. Instead, they suggest that verbatim memorization is the result of many interacting factors related to data and language modeling. Thus, removing verbatim memorized information without degrading model quality will be very difficult, especially for our best models.

\section{Related Work}

\paragraph{Verbatim memorization} LLMs can generate long sequences that are exact matches of their training data \cite{carlini2021extracting,carlini2023quantifying}.  Data repetitions, model size, and context length have been identified as factors correlated with verbatim memorization, mostly through observational studies~\cite{carlini2023quantifying,karamolegkou-etal-2023-copyright, prashanth2024recitereconstruct}. However, the mechanisms behind these factors are still not well understood. 
Prevention measures like fine-tuning, pruning, or string matching often degrade model quality or fail to cover variations of memorized sequences ~\cite{stoehr2024localizing,chang2024localization,ippolito-etal-2023-preventing}. This has motivated attempts to predict memorization before training~\cite{biderman2023emergent}. We revisit these findings and show that the challenges in prevention are caused by the entanglement between verbatim memorization and general language modeling.

\paragraph{Memorization and generalization} Memorizing individual examples is linked to generalization in deep neural models~\cite{arpit2017closer,zhang2017understanding}. Studies on image classifiers show memorizing noisy labels helps long-tail generalization~\cite{feldman2020longtail,feldman2020require}. 
However, verbatim memorization in LLMs differs from these settings, as texts memorized by LLMs are neither long-tail nor noisy. Recent work shows LLMs can memorize data without overfitting~\cite{tirumala2022memorization} and memorization of irrelevant information is necessary for accuracy~\cite{brown2021necessary}. We investigate the memorization--generalization relationship from the other direction, namely, how generalizable features learned from language modeling play a role in verbatim memorization.

\paragraph{Interpreting memory structures in Transformers}
MLP layers have been identified as key--value stores for structured knowledge and n-grams~\cite{geva-etal-2021-transformer,dai-etal-2022-knowledge,meng2022rome,geva-etal-2023-dissecting,voita2023neurons,haviv-etal-2023-understanding}, while other work has shown that attention heads can also store factual knowledge~\cite{allenzhu2024physics}. Yet, how the Transformer stores free-form texts has not been well studied, with mixed results that localize verbatim memorized texts to either MLP layers~\cite{chang2024localization} or attention heads~\cite{stoehr2024localizing}. In this work, we use causal intervention methods~\cite{pearl2001direct, pearl2009causality,beckers2019abstracting,vig2020gender,geiger2021causal}, which allow us to provide an approximate answer to the crucial counterfactual question of what would have happened if the model had not seen the memorized string%
~\cite{burg2021memorization,zhang2023counterfactual,feldman2020longtail,lesci2024causal}.

\section{A Framework for Studying Verbatim Memorization}\label{sec:framework}

We first introduce a framework to study the effects of language modeling quality on verbatim memorization in a tightly controlled setting. This framework adapts the data injection methods of \citet{jagielski2023measuring} and \citet{carlini2019secretsharer}, and aims to create minimally different models with and without specific sequences injected into their training data.

\paragraph{Sequence injection} 

We begin with a model checkpoint $\basemodel_{i}$. Let $O_{i}$ be the state of the optimizer at checkpoint $i$, and let $D_{i}$ be the final datapoint from the dataset $D$ that $\basemodel_{i}$ was trained on.
Using the state $(\basemodel_{i}, O_{i}, D_{i})$, we create two models. The \textbf{control model} $\controlmodel$ continues training $\basemodel_{i}$ for $s$ steps using the data $D_{[i:i+s]}$, with $O_{i}$ as the optimizer.  For the \textbf{treatment model} $\treatmodel$, we minimally alter $D_{[i:i+s]}$ to include a set of sequences $X$ that does not otherwise occur anywhere in $D$. Each sequence in this set is repeated uniformly every $m$ steps from a random offset, replacing the sequence at that point in $D$, until training step $i+s$.

The framework allows us to independently control three factors: the language model quality of $\basemodel$, the sequences $X$ to be memorized, and the frequency of the target sequence in the training data. Moreover, it creates approximate counterfactuals that allow us to observe what the model would be like if the model had not seen a particular sequence.

\paragraph{Optimizer state $O_{i}$} 

To simulate pre-training, we want an optimizer state that reflects the pre-training process prior to step $i$. Resetting the optimizer would lead to the first few batches
having an unduly large impact on the model loss. To achieve this, we first continue training the model $\basemodel_{i-t}$ from the pre-training checkpoint at $i-t$ over examples correspond to the next $t$ steps, using a freshly initialized optimizer. We then use the optimizer state of $\basemodel_{i-t}$ as the optimizer state $O_{i}$.

\paragraph{Measuring verbatim memorization} 

We define the \textit{verbatim memorization length} of an injected sequence $\seq$ as the number of tokens in the longest memorized substring, computed as follows: We prompt the model with all substrings of length 8, 16, 32, and 64 tokens extracted from $\seq$. We exclude prompts where the first 8 tokens of the continuation in $\seq$ is a substring in the prompt. For each prompt, we greedy decode the next 64 tokens as the prediction. Among all the predictions, we compute the longest prefix match between the prediction and the actual continuation in $\seq$ as the verbatim memorization length.

\section{Experiments}

We now report on a sequence of experiments aimed at helping to characterize the nature of verbatim memorization via the following four analyses.\footnote{Data and code available at \url{https://github.com/explanare/verbatim-memorization}}

\subsection{General Experimental Setup}
\label{sec:freq}

\paragraph{Models} We use checkpoints from the Pythia \texttt{160m}, \texttt{2.8b}, and \texttt{6.9b} models~\citep{biderman2023pythia} trained on the Pile~\cite{pile} deduped data.

\paragraph{Injection sequences} We curate a set of 100 sequences, each with 256 tokens, sampled from internet content published after the Pile cutoff date. We verify that the overlap between each sequence and the Pile is less than 50 characters (see Appendix~\ref{appx:inject_data}). Additionally, we create a set of 100 shuffled sequences by randomly shuffling tokens in each original sequence. The shuffled set preserves the overall vocabulary distribution but with little or no predictable structure.

\paragraph{Realistic injection frequencies} To determine realistic frequencies, we study the frequency range that triggers memorization using 5K sequence samples. A sequence is considered as memorized if it has a verbatim memorization length of 32 given a prefix of at most 32 tokens, We then hand-select a frequency where the \texttt{160m} model produces a mix of the memorized and non-memorized sequences, which is about every 10K to 100K examples. We detail the sampling and counting procedure in Appendix~\ref{appx:sequence_frequency}. Additionally, we observe that at the 6.9B scale, 94\% of memorized sequences occur at least 1~in~5M examples, which raises the question of whether a model could memorize a sequence it has seen only once. We address this question in \S\ref{sec:single_shot_illusion}, finding that purported instances are likely illusory.
\paragraph{Optimizer state} In \S\ref{sec:single_shot_illusion}, we use a freshly initialized AdamW optimizer~\cite{loshchilov2018decoupled}. In \S\ref{sec:checkpoint_vs_mem} and \S\ref{sec:ood}, we initialize the optimizer by pre-training on 1M examples.

We provide detailed setup in Appendix~\ref{appx:training_setup}.

\subsection{The Illusion of Single-shot Verbatim Memorization}
\label{sec:single_shot_illusion}

Do LLMs verbatim memorize a sequence in pre-training after only seeing the sequence once?

\begin{table*}
\begin{center}
\small
\begin{tabular}{c c c}
\toprule
 Type & Percentage & Description \\
 \midrule
Template & 54 & Templated texts, where the variable content is usually provided in the prompt  \\
Variation & 21 & Spacing, punctuation, and textual variants of texts  \\
Induction & 17 & Texts with inductive patterns, e.g., ordered number sequences or repeating sequences \\
Composition & 8 & Texts composed of frequent patterns, with composition rules specified by the prompt \\
\bottomrule
\end{tabular}
\vspace{-2ex}
\caption{Four types of sequences that create the illusion of single-shot verbatim memorization.}
\label{tab:single-occur-pattern}
\end{center}
\vspace{-4ex}
\end{table*}

We first manually annotate the 6\% low-frequency sequences verbatim memorized by the \texttt{6.9b} model in \S\ref{sec:freq} and identify four patterns that create the illusion of single-shot verbatim memorization in Table~\ref{tab:single-occur-pattern}, with examples shown in Appendix~\ref{appx:single_shot_illusion}. These seemingly low-frequency sequences are either under-counted due to limitations of string-based matching or simply not verbatim memorized, i.e., a checkpoint produced before the sequence occurs can already generate the sequence verbatim. \citet{prashanth2024recitereconstruct} has also identified similar sequences as ``Reconstruction'' and ``Recollection''. These patterns suggest not all tokens in the verbatim memorized sequences are actually memorized, some might be completed by the LM.

One may argue that a memorized sequence that only occurs once in the training data is inherently hard to discover. To complement counting, we directly measure a model's ability to verbatim memorize a sequence after one occurrence.

\paragraph{Setup} We train the \texttt{2.8b} and \texttt{6.9b} 80K checkpoint for 200 steps, where a sequence to memorize is injected into \emph{the first batch}. We measure the verbatim memorization length every 10 steps. 

\begin{figure}[t!]
    \centering
\vspace{-0.5ex}
\includegraphics[width=0.95\linewidth,trim={0 5 0 5},clip]{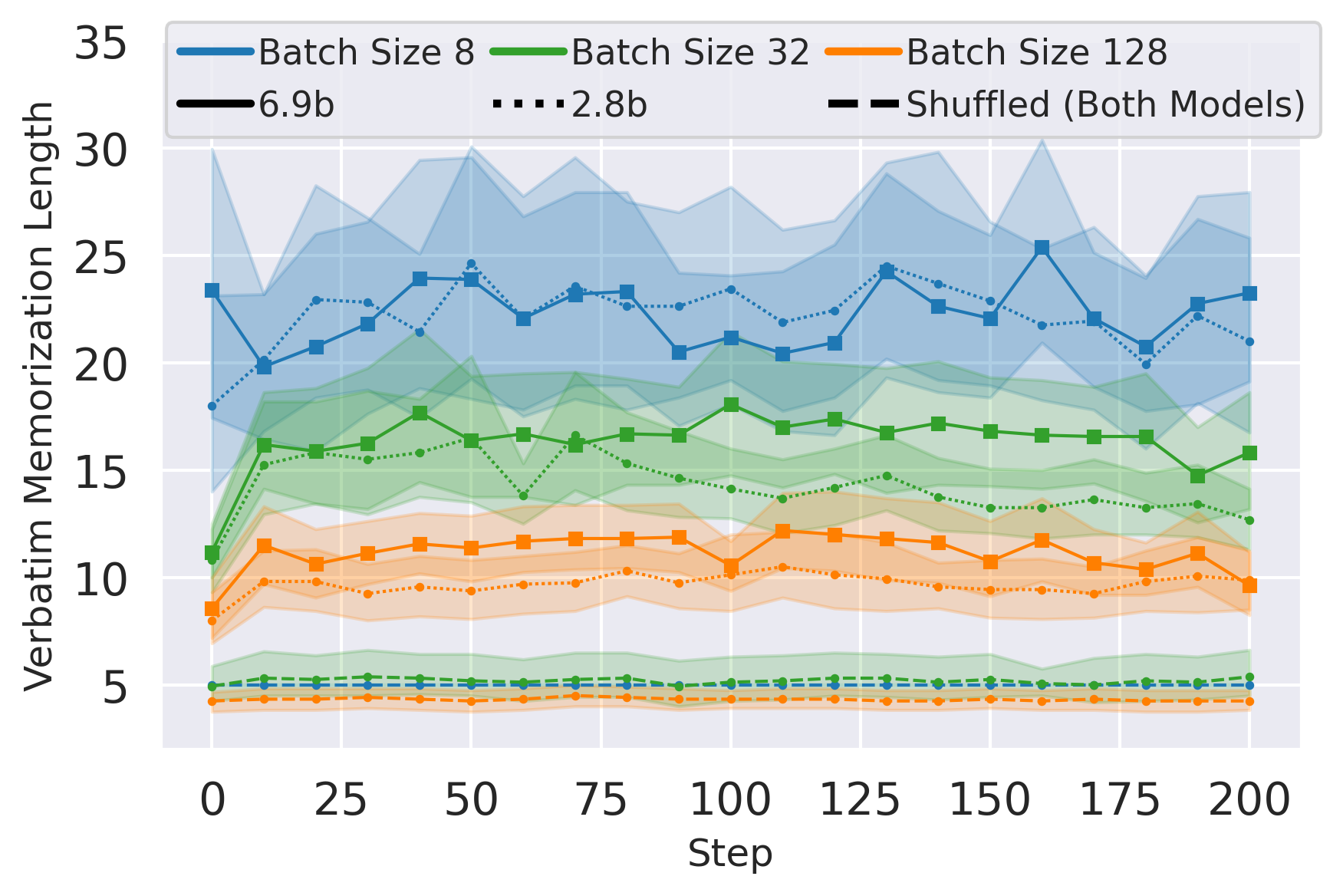}
\vspace{-2.5ex}
    \caption{Single-shot verbatim memorization length of the \texttt{2.8b} and \texttt{6.9b} models after 200 training steps.}
    \label{fig:exp-single-shot-vs-step}
    \vspace{-4ex}
\end{figure}

\paragraph{Results} Results are shown in Figure~\ref{fig:exp-single-shot-vs-step}, averaged over 16 injection sequences and their shuffled version. The verbatim memorization length decreases significantly as the batch size increases. Moreover, the verbatim memorization length peaks around 25--100 steps after seeing the injected sequence, likely due to momentum terms in the optimizer~\cite{chang2024largelanguagemodelsacquire}. Even at the peak, the \texttt{6.9b} model only verbatim memorizes 12$\pm$3.7 tokens from the original sequences at batch size 128. Shuffled sequences are memorized 5$\pm$1 tokens regardless of batch size or model size. With a batch size of 1024 in pre-training, it is extremely unlikely that models with a size smaller than 6.9B can just verbatim memorize an arbitrary sequence in a single-shot. 

\begin{figure}[t]
    \centering
    \vspace{-1ex}
   \includegraphics[width=1.0\linewidth,angle=0,trim={0 25 0 5},clip]{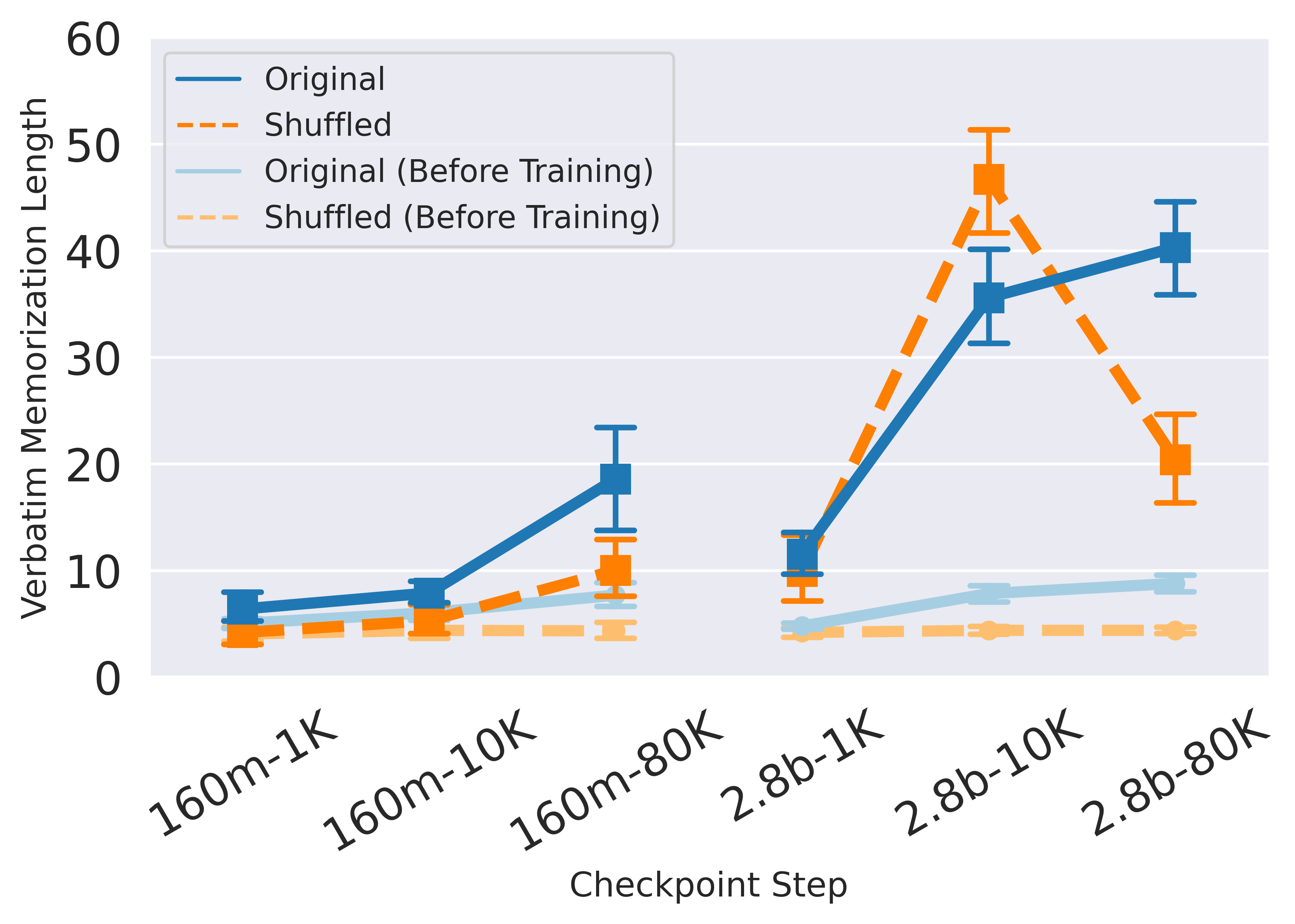}
     \vspace{-4ex}
     \caption{Pythia checkpoint vs.~verbatim memorization length of the original and shuffled sequences.}
    \label{fig:exp-ckpt-vs-mem}
    \vspace{-4ex}
\end{figure}

\subsection{Better LMs Memorize Longer Sequences}
\label{sec:checkpoint_vs_mem}

Are better LMs more likely to memorize sequences? Intuitively, better LMs are those that achieve lower perplexity on novel sequences drawn from their training distribution. From this perspective, we might expect them to be better at memorizing such sequences as well, since they simply require fewer bytes on average to encode such sequences~\cite{deletang2024language}. 

\paragraph{Setup} To decouple model quality from model size, we experiment with three checkpoints at 1K, 10K, and 80K steps from the \texttt{160m} and \texttt{2.8b} models.
We use two injection frequencies for both models: every 50K and 10K examples. For each model run, we pack a set of 4--10 sequences, so that the total amount of injected sequences is less than 0.04\% of the training data. We measure the verbatim memorization after 40 and 20 occurrences for the two models respectively, as with 20 occurrences the \texttt{2.8b} model can already memorize longer sequences than the \texttt{160m} one. 

\paragraph{Results} Figure~\ref{fig:exp-ckpt-vs-mem} (solid blue lines) shows the result of 1~in~50K frequency, with results of 10K frequency in Appendix~\ref{appx:ckpt_vs_mem}. The findings are clear: later checkpoints memorize more, and the larger model is able to memorize more, even when seeing the sequences fewer times. Overall, checkpoints corresponding to higher quality models are more likely to memorize the injected sequences.

\subsection{OOD Sequences Are Harder to Memorize}\label{sec:ood}

The previous section shows that better models are more capable of memorizing sequences in their training distributions. What about sequences from a different distribution?
One hypothesis is that out-of-domain sequences are more likely to be memorized, as they contain rare sequences of tokens that can be used as identifiers to recall the memorized content~\cite{tirumala2022memorization}.
The other hypothesis is that in-domain sequences are more likely to be memorized, because they have lower perplexity before training, as in the single-shot case in \S\ref{sec:single_shot_illusion}.

\paragraph{Setup} To investigate which hypothesis holds, we use the set of shuffled sequences, which naturally have a higher perplexity than the original sequences when measured using the model $M$, i.e., before training on the sequences. We follow the training and evaluation protocol from \S\ref{sec:checkpoint_vs_mem}.

\paragraph{Results} Results are shown in Figure~\ref{fig:exp-ckpt-vs-mem} (dashed orange lines). On average, the original sequences drawn from the training distribution are more likely to be verbatim memorized than shuffled sequences, except for the \texttt{2.8b-10K} checkpoint. Even though the shuffled sequences are not memorized as well, we do still see the trend in which later checkpoints memorize longer sequences. In terms of perplexity changes, the perplexity of memorized shuffled sequences does decrease faster during training than the perplexity of original sequences.

These findings suggest that the verbatim memorization observed in pre-trained LLMs is more complex than recalling a sequence based on some unique identifiers, as otherwise we would see the models memorize more shuffled sequences. Multiple factors might contribute to the process: general mechanisms to efficiently  store free-form texts, as well as structures that favor in-domain sequences. The former might be learned from modeling random sequences such as URLs, UUID, or even digits of $\pi$, while the later might emerge as modeling structures in natural languages. We investigate these mechanisms in the following sections.

\subsection{Memorization is Triggered by Abstract Model States Distributed Across Tokens}\label{sec:trigger_dependency}

A core question for verbatim memorization is how models encode memorized sequences. We consider two aspects of the question: (1) Which tokens encode the information of the verbatim memorized sequences? (2) Do models encode token-level information (low-level representations) or more abstract states of the model (high-level representations)?

To answer these questions, we seek to identify the causal connections between the sequence that triggers memorization and the tokens in the verbatim memorized sequence. In more detail,
consider a treatment model $\treatmodel$ that takes as input a \emph{trigger} prefix $\textbf{p} = x_{1}, \dots x_{n}$ and output a 
verbatim memorized sequence $\seq = x_{n+1}, \dots x_{n+k}$. From this it follows that the trigger prefix $\textbf{p}$ creates an internal state $\mathbf{s}$ in $\treatmodel$ that is sufficient for generating~$\seq$. %
If verbatim memorized information is \emph{localized} to the trigger $\textbf{p}$, then \emph{every} token in $\seq$ should have a causal connection to the internal state $\textbf{s}$.

\paragraph{Interventions} 
To test whether every token in $\seq$ in fact depends on $\textbf{s}$, 
we use \emph{interchange interventions} (also known as activation patching;~\citealt{geiger2021causal,vig2020gender}), which is calculated as follows.  %
First, let $\texttt{GetVal}(M(x), l)$ be the representation $\mathbf{v}$ that model $M$ computes at location $l$ when it processes input $\mathbf{x}$. Second, let $M_{l \gets \mathbf{w}}(x)$ be the model that is just like $M(x)$ except that the representations computed at location $l$ have been replaced by the values $\mathbf{w}$. An interchange intervention is one in which the value used in this intervention is one created when the model processes a different input $x'$. This results in a nesting of $\texttt{GetVal}$ inside the intervention:
$M_{l \gets \texttt{GetVal}(M(x'), l)}(x)$.
In other words, the interchange intervention replaces the values computed at $l$ with those obtained from processing a different example. 

\begin{figure*}[ht]
    \small
    \centering
    \begin{subfigure}[t]{0.33\linewidth}
\includegraphics[align=t,width=\linewidth,trim={0 2.3cm 14cm 1.5cm},clip]{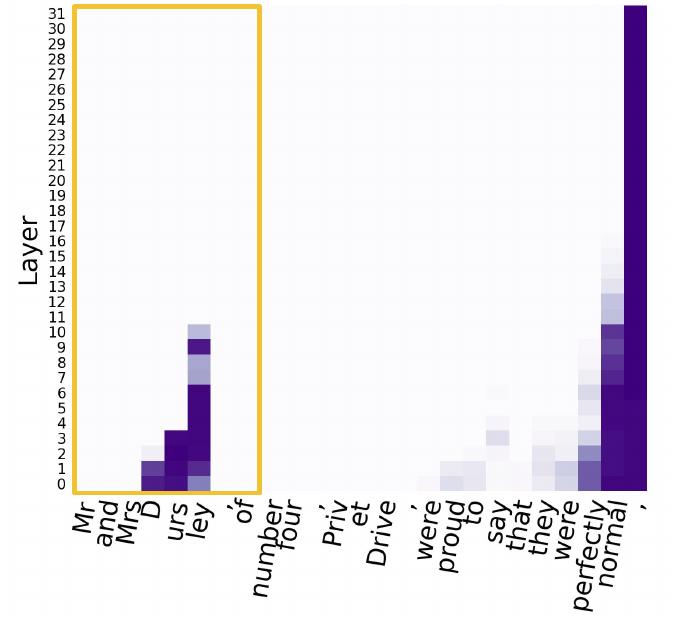}
     \vspace{-3ex}
    \caption{}
    \label{fig:causal-dependencies-thank}
    \end{subfigure}%
    \hfill
    \begin{subfigure}[t]{0.30\linewidth}
\includegraphics[align=t,width=\linewidth,trim={1.1cm 2.3cm 14cm 1.6cm},clip]{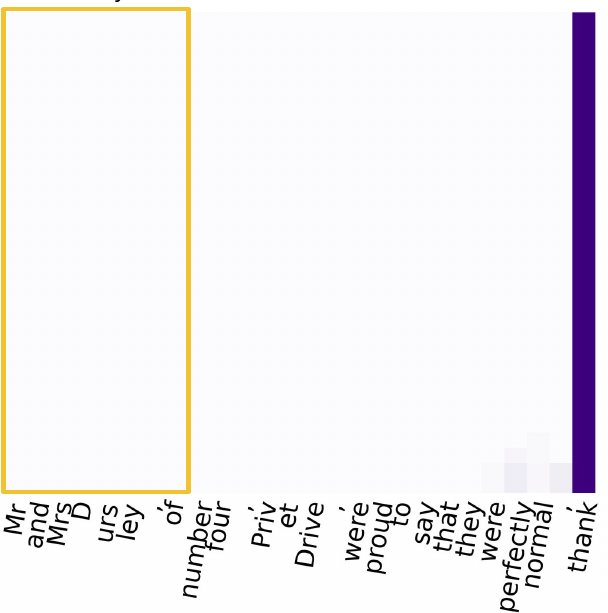}
    \vspace{-3ex}
    \caption{}
   \label{fig:causal-dependencies-you}
    \end{subfigure}%
    \hfill
    \begin{subfigure}[t]{0.36\linewidth}
    \begin{subfigure}[t]{0.98\linewidth}
\includegraphics[align=t,width=\linewidth,trim={0 0 10 0},clip]{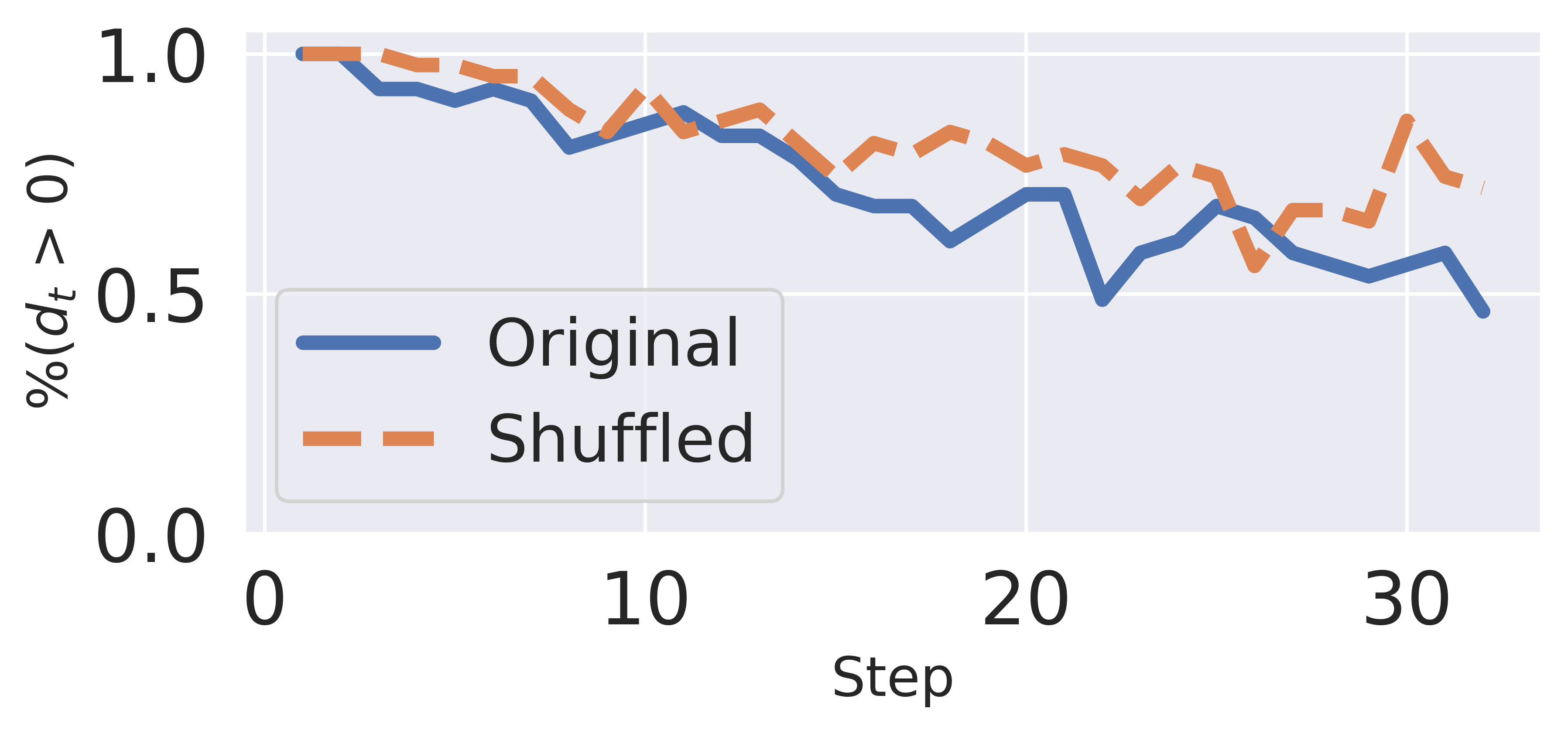}
    \vspace{-4.5ex}
    \caption{\phantom{c}}
    \label{fig:dep-vs-step}
    \vspace{-0.5ex}
    \end{subfigure}
    \begin{subfigure}[t]{0.98\linewidth}
\includegraphics[align=t,width=\linewidth,trim={0 0 10 15},clip]{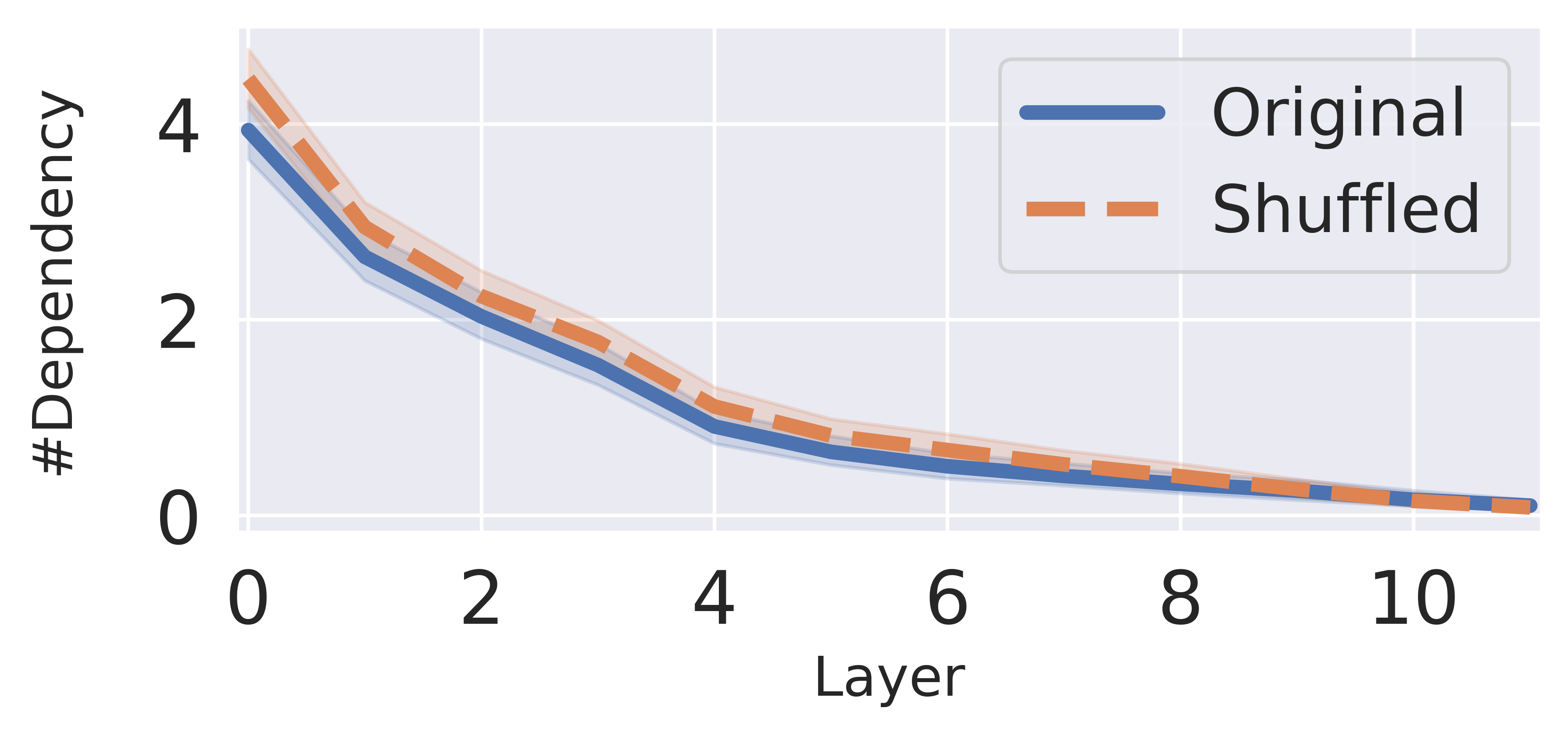}
    \vspace{-4.2ex}
    \caption{\phantom{d}}
    \label{fig:dep-vs-layer}
    \end{subfigure}%
    \end{subfigure}%
    \vspace{-2.8ex}
    \caption{Causal dependencies between the trigger and verbatim memorized tokens. (a) An example of a memorized token that depends on the trigger (the yellow box). The example is the first sentence of the book \textit{Harry Potter and the Philosopher's Stone}, which is in the Pile. (b) An example of a memorized token that does not depend on the same trigger. (c) The percentage of memorized tokens that causally depend on the trigger decreases by step. (d) For memorized tokens that depend on the trigger, there is on average one causal dependency even at the middle layers.}
    \label{fig:exp-causal-dependencies}
    \vspace{-3.5ex}
\end{figure*}

Our interchange intervention focuses on the decoding process given a trigger prefix. Suppose $\treatmodel$ has run a forward pass on $\mathbf{p} = x_1, \dots, x_{n+1}, \dots, x_{n+t-1}$ and so is going to predict token $t$. Our interchange intervention 
replaces the residual stream representations in layer $k$ of a token $x_{j}$ (for $j \leq n$) in $\mathbf{p}$ with residual stream representations extracted at the same layer and token position where the model input is a random sequence $\mathbf{r}$ sampled from the Pile:
\begin{multline}
\treatmodel_{(j,k) \gets \mathbf{v}}(\mathbf{x}) \\
\text{ where } \mathbf{v} = \texttt{GetVal}(\treatmodel(\mathbf{r}), (j,k))
\end{multline}
If the next token prediction is causally dependent on the trigger representation, we expect the predicted token to change after this intervention, since the chance of a random sampled token having a similar representation as the intervened token is extremely low. Alternatively, if the representation has no causal effect, we expect the output to be the same. Interventions on model representations allow us to measure which tokens have causal effects (our question 1 above), and at which layers the token information was used by the model to decode the next memorized token (our question 2).

\paragraph{Metrics} Let $L$ be the set of all possible intervention locations, i.e., the set of residual streams across all trigger token position and all layers. Let $p_{l}$ represent the percentage of interventions that lead the model to output the verbatim memorized token when intervened on at~$l$. We estimate the causal effect of the trigger on the verbatim memorized token at step $t$ as a causal dependency score:
\begin{equation}
   d_{t}=1 - \max_{l\in L}\{p_{l}\} 
\end{equation}
The score is between 0 and 1, where $1$ means strong causal dependency on the trigger and $0$ means no causal dependency. By definition, $d_1=1$, since the last layer residual stream at the last token always has causal effects on the first predicted token.

We define the number of dependencies a memorized token has over a set of locations $L$ as $N_{t}$:
\begin{equation}
   N_{t}=\sum_{l\in L}\mathbbm{1} [p_{l} > T]
\end{equation}
where $T$ is a threshold that we set to 0.1 to filter dependencies with weak causal effects.

\paragraph{Setup} We analyze the  models in \S\ref{sec:checkpoint_vs_mem} trained from the \texttt{160m-80K} checkpoint with an injection frequency of 1 in 10M examples. We sample 50 injected sequences, including both original and shuffled ones. We additionally verified the patterns generalize to the \texttt{6.9b} models using 50 memorized sequences sampled from the 5K sequences in \S\ref{sec:freq}.

\paragraph{Results} Figure~\ref{fig:exp-causal-dependencies} summarizes our results: (1) Not all tokens in the verbatim memorized sequence are causally dependent on the trigger representations, e.g., Figure~\ref{fig:causal-dependencies-you}.
Instead, these tokens often exhibit dependencies that resemble syntactic structures, e.g., the direct object depends on the preceding verb and the closing parenthesis depend on the opening parenthesis. For sequences with no clear structure, memorized tokens depend on more trigger tokens that are relatively rare, a pattern observed in previous work~\cite{tirumala2022memorization,stoehr2024localizing}.
(2) Most memorized tokens depend on higher-level representations produced by middle layers. At layer 4, there still exists on average one dependency. We observe similar  patterns in the \texttt{6.9b} model (see Appendix~\ref{appx:dependency}).
Our results show that information about the verbatim memorized sequence is (1) distributed across tokens and (2) encoded in abstract states as opposed to token-level features. There is simply \emph{no} representation of the trigger $\textbf{p}$ that causally encodes the \emph{entire} memorized sequence.

Moreover, the fact that not all verbatim memorized tokens are causally dependent on the trigger suggests the model might only memorize information about a subset of tokens and fill in the gaps with the general LM. The verbatim memorized sequence might be \emph{reconstructed} token-by-token, where each token is predicted using different mechanisms depending on the structures involved.
This might explain why in-domain sequences are more likely to be memorized. In fact, the two mechanisms we observed -- attending to syntactic structures and rare tokens -- are identified in Transformers that have not seen or memorized a particular sequence~\cite{tian2023scan,chen2024sudden}. 
Lastly, models encode abstract states as opposed to token-level information, which might explain why the memorized sequences can be triggered in a context that is different from the one seen in training. We further test this hypothesis in \S\ref{sec:cross_model_intervention}.

\subsection{Verbatim Memorization Leverages General Language Modeling Capabilities}
\label{sec:cross_model_intervention}

The results of \S\ref{sec:checkpoint_vs_mem} and~\S\ref{sec:ood} provide behavioral evidence that memorization depends on general language capabilities. In this section, we extend the intervention-based methods of \S\ref{sec:trigger_dependency} in an effort to  characterize this relationship in terms of the underlying computation they share. The core analytic technique is an interchange intervention that seeks to get a control model $\controlmodel$ to produce memorized strings by intervening with internal states from a minimally different treatment model $\treatmodel$.

Our core finding is that, while such interventions do not lead $\controlmodel$ to produce entire memorized sequences, we can get it to produce the first few tokens of such sequences. %
Moreover, among the interventions at a layer that do produce memorized tokens, more than 50\% can still produce the same memorized tokens using model components at the corresponding layer from $\controlmodel$, which are weights learned only from general language modeling.

\paragraph{Cross-model interchange interventions} 

We propose a novel intervention that replaces representations in $\controlmodel$ with corresponding ones in $\treatmodel$:
\begin{equation}
  \controlmodel_{l\leftarrow \textbf{v}}(\mathbf{p}) 
  \text{ where } \textbf{v} = \texttt{GetVal}(\treatmodel, \mathbf{p} , l)
\end{equation}
Suppose $\mathbf{p}$ is a triggering prefix for a memorized generation $\seq$. If this intervention leads $\controlmodel_{l\leftarrow \textbf{v}}(\mathbf{p})$ to generate parts of $\seq$, then we have evidence that the memorization behavior was guided in part by the representation at $l$ and in part by the general structure of $\controlmodel$.

It may seem surprising to transfer representations between two models. However, the models begin from the same checkpoint and are trained on almost identical sequences. This weakly suggests that their representations will be compatible. In addition, prior work has shown that even representations from different families of models are interchangeable with some affine transformations \cite{ranzato2021similarity,ghandeharioun2024patchscopes}. We also experimentally verify the coherence of these interventions in our results section below.

As in \S\ref{sec:trigger_dependency}, we explore intervention sites across all layers, since we do not know a priori where the relevant information might be stored. For each layer, we target both attention and MLP components, which have been identified as related to memorization behaviors in Transformer-based LMs ~\cite{geva-etal-2021-transformer,dai-etal-2022-knowledge,geva-etal-2023-dissecting,stoehr2024localizing,allenzhu2024physics}. We aim to understand to what extent these components  reuse computation learned from general language modeling.

\begin{figure}[!t]
    \centering
    \includegraphics[width=\linewidth,trim={0 8.5cm 10cm 0},clip]
    {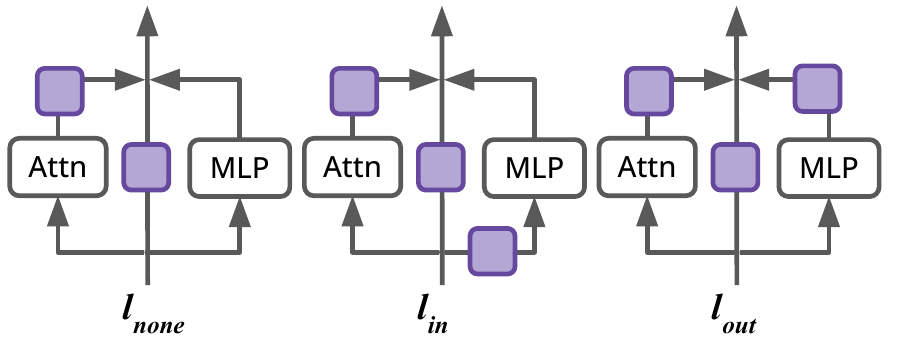}
       \vspace{-4ex}
    \caption{Three sets of cross-model interchange interventions that allow us to measure to what extent models reuse components learned from general language modeling in verbatim memorization.} 
    \label{fig:cross_intervention}
    \vspace{-4ex}
\end{figure}

\paragraph{Metrics} For an intervention at location $l$, let $p_{l,n}$ be the percentage of examples where the first $n$ tokens predicted by $\controlmodel$ match the verbatim memorized tokens generated by $\treatmodel$. We consider small values of $n\in [1, 2, 4]$; as we will see, by $n = 4$, success rates have gone to effectively $0$.

For each layer, we want to measure whether verbatim memorization reuses computation learned from general modeling, i.e, computation defined by components in $\controlmodel$. We compute $p_{l,n}$ at three sets of intervention locations across all trigger tokens.
We use MLP as an example in Figure~\ref{fig:cross_intervention}.
\begin{itemize}\setlength{\itemsep}{0pt}
    \item $l_{\textit{none}, i}$: Attention output at layer $i$ + Residuals at layer $i-1$
    \item $l_{\textit{in}, i}$: $l_{\textit{none}, i}$ + MLP \textbf{input} at layer $i$
    \item $l_{\textit{out}, i}$: $l_{\textit{none}, i}$ + MLP \textbf{output} at layer $i$ (i.e., Residuals at layer $i$)
\end{itemize}
In $l_{\textit{none}, i}$, the residual from the treatment model $\treatmodel$ is not propagated into the MLP layer of the control model $\controlmodel$. The MLP output is still computed using the MLP input from $\controlmodel$. The locations for attention can be defined symmetrically.

Let $R_{i,n}$ be the percentage of interventions that lead to $\controlmodel$ producing a memorized short prefix of $n$ tokens using only MLP input from the treatment model $\treatmodel$, but not the MLP layer weights from the treatment model $\treatmodel$:
\begin{equation}
   R_{i, n}=\frac{p_{l_{\textit{in}, i}, n} - p_{l_{\textit{none}, i}, n}}{p_{l_{\textit{out}, i}, n} - p_{l_{\textit{none}, i}, n}} 
\end{equation}
$R_{i,n}$ is only meaningful when the denominator is sufficiently large, i.e., the layer has causal effects on the next $n$ verbatim memorized tokes. A higher $R_{i,n}$ value suggests that the MLP (or attention) component in $\controlmodel$ plays a similar causal role on the next $n$ memorized tokens as the corresponding component in $\treatmodel$. In other words, a sign of leveraging general language modeling capabilities. %

\paragraph{Setup} We use the $\texttt{160m}$ models in \S\ref{sec:checkpoint_vs_mem} trained from the step 80K checkpoint, with treatment model data injected at a frequency of every 10K examples. We analyze 2,000 tokens predicted as part of 120 verbatim memorized sequences (including shuffled sequences), which covers about 1000 distinct tokens. About 25\% of these verbatim memorized tokens can be correctly predicted by the control model as well. We exclude these from further analysis. However, these tokens suggest that a quarter of the verbatim memorized tokens result from general language modeling. For the remaining tokens, we compute $p_{l_{out, i}, n}$ and $R_{i, n}$.

\begin{figure}[t]
    \centering
   \includegraphics[width=0.8\linewidth,trim={0 0 0 5},clip]{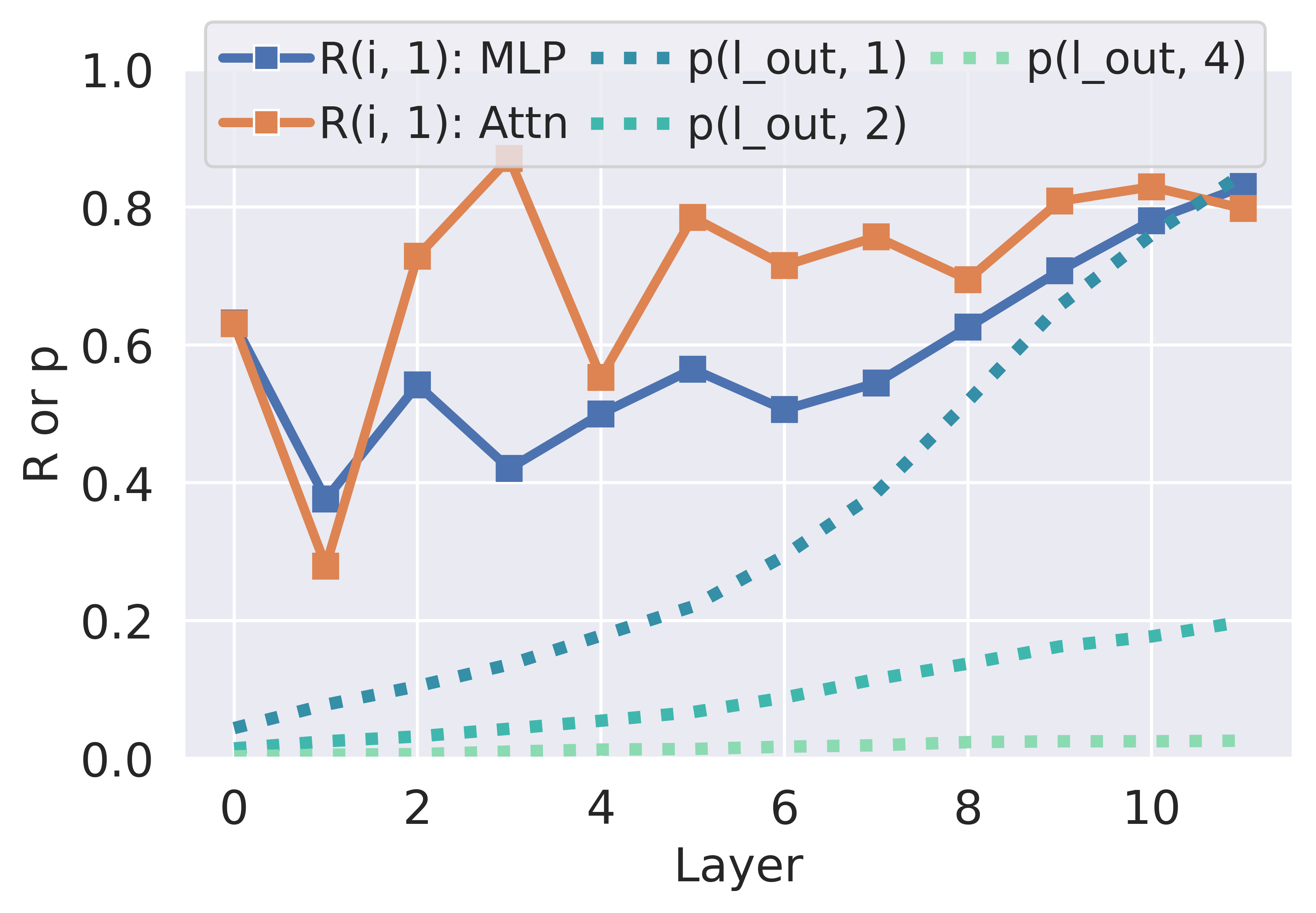}
   \vspace{-2ex}
     \caption{Results of cross-model interventions. Dotted lines: There are interventions that can control $\controlmodel$ to produce the next 1--2 memorized tokens, but not any longer. Solid lines: Among interventions that produce the next memorized token, more than 50\% can still produce the same token using components of $\controlmodel$.}
    \label{fig:exp-cross-model-intervention}
    \vspace{-4ex}
\end{figure}

\paragraph{Results} Results are shown in Figure~\ref{fig:exp-cross-model-intervention}. We first look at dotted lines: (1) When intervening on the last layer residuals, $p_{l_{out, 11}, n}=85$\%, which validates our intervention setup -- representations from the two models are indeed interchangeable for the majority of the inputs. (2) As $n$ increases,  $p_{l_{out, i}}$ drops to almost zero, which means interventions on individual model components have little to no causal effect on producing the memorized prefixes. This aligns with our findings in \S\ref{sec:trigger_dependency}: Memorized information is distributed across tokens.

For the solid lines, which are $R_{i, 1}$, a setting where a significant percentage of interventions can produce memorized tokens, we find that (1) the majority of attention and MLP layers have a $R_{i,1}$ values above 50\%, suggesting the $\treatmodel$ model is performing similar computation as $\controlmodel$, which are computations learned from general language modeling. In fact, verbatim memorization can still happen with frozen attention heads, i.e., only using attention patterns learned from general language modeling (Appendix~\ref{appx:frozen_components}). (2) There are a few layers where $R_{i,1}$ is around 30\%, i.e., MLP components in layer 1 and 3 and attention in layer 1 and 4, suggestion these components are largely different between $\controlmodel$ and $\treatmodel$ and likely store memorized information. Indeed, previous work that uses gradient-based approaches also indicates that lower layers play an important role in verbatim memorization~\cite{stoehr2024localizing}. However, an $R_{i,1}$ around 30\% means it is still challenging to fully isolate the memorized information, even just for predicting a single token.

\paragraph{Analysis} The ability to leverage computations learned from general language modeling provides an explanation of why higher quality models verbatim memorize more sequences. This also suggests that verbatim memorization is fundamentally intertwined with language modeling capabilities, as the control and treatment models largely share both attention and MLP structures across multiple layers. 

\section{Stress Testing on Unlearning Verbatim Memorized Texts}
\label{sec:stress_testing}

Given the nature of the verbatim memorization discussed in \S\ref{sec:trigger_dependency} and \S\ref{sec:cross_model_intervention}, we propose a suite of automated stress tests to evaluate whether unlearning methods truly remove verbatim memorized information without systematically altering the LM.

\subsection{A Stress Testing Dataset}

Our stress tests are built on two observations:
\begin{itemize}\setlength{\itemsep}{0pt}
    \item Memorized information is distributed across tokens, hence evaluation should include prompts that cover different spans of a memorized sequence (``Position Perturbations'').
    \item Verbatim memorization is triggered by abstract model states, hence evaluations should cover semantically similar variations of the prompt trained on (``Semantic Perturbations'').
\end{itemize}

Consider a trigger prompt of $n$ tokens $x_1\dots x_n$ with memorized continuation $x_{n+1}\dots x_{n+k}$ in the original training set (which is also the evaluation set in the unlearning setup). For ``Position Perturbations'', we generate two sets of perturbed prompts: 
\[
\{x_{1}\dots x_{n+i}| i \in [0, t]\} \cup \{x_{n-i}\dots x_{n}| i \in [t, n)\}
\]
For ``Semantic Perturbations'', we replace each word or a consecutive sequence of digits (or characters) in the prompt with a similar word.
\[
\{x_{1} \dots s_{i} \dots x_{n}| i\in[1, n], s_{i}\in \mathcal{S}(x_{i})\}
\]
where $\mathcal{S}(x_{i})$ is a set of similar words of $x_{i}$. Example stress tests are in Appendix~\ref{appx:unlearning}.

\subsection{Evaluation}

We evaluate the gradient ascent, sparse fine-tuning, and pruning methods proposed in \citet{stoehr2024localizing,chang2024localization}. %
These methods have been shown to prevent models from generating verbatim memorized texts on the fine-tuned prompts, at the cost of increasing perplexity on other texts \cite{chang2024localization,stoehr2024localizing}.

\paragraph{Setup} We follow the setup in \citet{stoehr2024localizing} (see Appendix~\ref{appx:stree_test_setup}). Given a 50-token trigger prompt and a 50-token continuation memorized by the GPT-Neo 125M model, the goal is to unlearn the continuation while retaining model quality on other prompts. For each sequence, we generate $\sim$1K perturbed prompts with $t=20$ for Position Perturbations and use ChatGPT to generate around 10 similar word substitutions per word for Semantic Perturbations. For both original and stress test prompts, we report the longest continuation that matches the memorized sequence. For stress test prompts, the length is max-pooled over all prompts.

\begin{table}
\begin{center}
\resizebox{1\linewidth}{!}{
\begin{tabular}{@{} c  c  c  c @{}}
\toprule
 Method & Original & Position & Semantic \\
 \midrule
Gradient Ascent & 19$\pm$18 & 35$\pm$15 & 31$\pm$21\\
Sparse Fine-tune & 23$\pm$20 & 36$\pm$16 & 34$\pm$21 \\
Neuron Pruning & 4$\pm$4 & 14$\pm$8 & 11$\pm$10 \\
\bottomrule
\end{tabular}
}
\caption{The exact match length of model outputs with the original and stress testing prompts. On average, stress testing prompts can extract 10--15 more tokens.}
\vspace{-3ex}
\label{tab:stress-test-unlearning}
\end{center}
\end{table}

\paragraph{Results} Table~\ref{tab:stress-test-unlearning} shows the results, with full length distributions shown in Appendix~\ref{appx:unlearning}. On average, the perturbed prompts increase the exact match length by 10--15 tokens. For the gradient ascent and sparse fine-tuning, the stress tests increase the fully extractable sequences, i.e., exact match of  50 tokens, from 22\% to 56\%. The neuron pruning method is more robust to the stress tests, however, it often leads to degeneration on the perturbed prefixes, e.g., outputting repetitive texts. Overall, while these unlearning methods largely prevent models from generating the verbatim memorized sequence given a particular prefix, they do not completely remove the verbatim memorized information -- the model can still generate the memorized texts when prompted with variants of the prefix.

\section{Discussion and Conclusion}
\label{sec:conclusion}
Verbatim memorization is a pressing issue for LM research, as it has ramifications for privacy, copyright, and other legal issues. Thus, one might hope that we will find ways to identify and control memorization.
The present paper suggests that such control may be extremely difficult to achieve because verbatim memorization is thoroughly intertwined with general language modeling quality. For example, given current training procedures, LMs will memorize more strings as their quality improves. Strings that resemble those from the LM's training data are more likely to be memorized (\S\ref{sec:checkpoint_vs_mem}), but even OOD strings (which may include private identifiers, usual metadata patterns, etc.)\ are memorized at non-trivial rates by our best models (\S\ref{sec:ood}).

In light of these findings, one might simply accept that LMs will memorize strings and try to mitigate memorization by blocking specific triggering strings. Unfortunately, this method is bound to have very low recall. As we showed in \S\ref{sec:trigger_dependency}, the notion of a trigger is extremely complex. Overall, the trigger is actually a set of distributed model-internal states that encode generalizable high-level features that numerous inputs can lead to. In \S\ref{sec:cross_model_intervention}, we deepened this result by showing that even a control model that has never seen a specific memorized input $\seq$ can be made to produce parts of $\seq$ via an intervention from a model that has memorized $\seq$. In \S\ref{sec:stress_testing}, we show the practical implications of these distributed, abstract triggering states on unlearning methods, which lead to failures in removing verbatim memorized information or degrading general model quality. These results all point to the idea that generating memorized strings is in part simply language model decoding as usual.

More broadly, these findings suggest that ``verbatim memorization'' is something of a misnomer, as the phenomenon involves memorization of more abstract model states as opposed to only memorization of token-level information. Thus, to be successful, future attempts to control memorization will likely require new techniques for characterizing and controlling these abstract model states. Such techniques are likely to greatly improve our understanding of LLMs in general.

\section*{Limitations}

Our work contributes to understanding verbatim memorization behaviors in LLMs, an important problem that has practical implications and applies to almost all LLMs trained on large-scale web corpora. However, constrained by the availability of fully open sourced LLMs (i.e., LLMs with training dataset, checkpoints, and training hyperparameters fully available), we only conducted experiment on the Pythia family of models, focusing on model sizes up to \texttt{2.8b}. As more fully open sourced models come out, such as OLMo,\footnote{\url{https://allenai.org/olmo}} we would like to see if our findings on Pythia models generalize to other model families.

One important finding of our paper is that verbatim memorization actually involves memorization of abstract model states as opposed to just token-level information. This raises the concern of whether focusing on verbatim memorization reveals the full scale of what models actually memorize. LLMs could memorize long sequences of abstract states as well, which might remain undetected if we only focus on verbatim memorization. For example, models memorize syntactic templates~\cite{shaib2024templatemem}. We discuss these findings in \S\ref{sec:conclusion}.

\section*{Acknowledgments}

We would like to thank Atticus Geiger, Jiaao Chen, Robin Jia, John Hewitt, Ken Liu, Zhengxuan Wu, and Aryaman Arora for insightful discussion and feedback.  This work is supported in part by a grant from ONR and a research grant from Sandia National Laboratories. Any opinions, findings, and conclusions or recommendations expressed in this material are those of the authors and do not necessarily reflect the views of Sandia National Laboratories.

\bibliography{anthology,custom}

\appendix
\onecolumn

\section{Sequence Frequency vs. Verbatim Memorization}
\label{appx:sequence_frequency}

\subsection{Choice of Sequence Injection Frequencies}
To estimate a realistic frequency for sequence injection, we need to know roughly what percentage of sequences in the Pile are memorized at each frequency range. The deduped version of Pile contains about 98M sequences, each of length 2048 tokens. Ideally, one would build an index of the entire corpus to count substrings, as is done in~\citet{carlini2023quantifying,liu2024infinigram}. However, the storage and computation required to build an index is costly. We employ a sampling-based approach instead.

\paragraph{Sampling} We first describe how to sample a relatively small set of sequences to estimate memorization rates at each frequency range. We start with random sampling 1M sequences of length 128 from the Pile and compute verbatim memorization length using the \pythia{6.9b} model. For a sequence to be considered memorized, the sequence must have a verbatim memorization length of at least 32, (i.e., there must exist a substring of length $\leq$ 32 tokens, such that when prompted with this substring, the model outputs a continuation that matches the next 32 tokens or more). Among the 1M sequences, there are about 9K memorized sequences and 991K non-memorized sequences. Next, we randomly sample 2.5K memorized and 2.5K non-memorized sequences, which means that we downsample the non-memorized sequences 110 times relative to memorized ones. For each sequence, we further sample a substring of 16 tokens. For memorized sequences, the 16 tokens are sampled from the memorized substring, i.e., the model outputs instead of the prompts. We refer to these 5K 16-token sequences as probes.

\begin{figure*}[ht]
    \small
    \centering
    \begin{subfigure}[t]{0.32\linewidth}
\includegraphics[width=\linewidth,trim={5 0 0 20},clip]{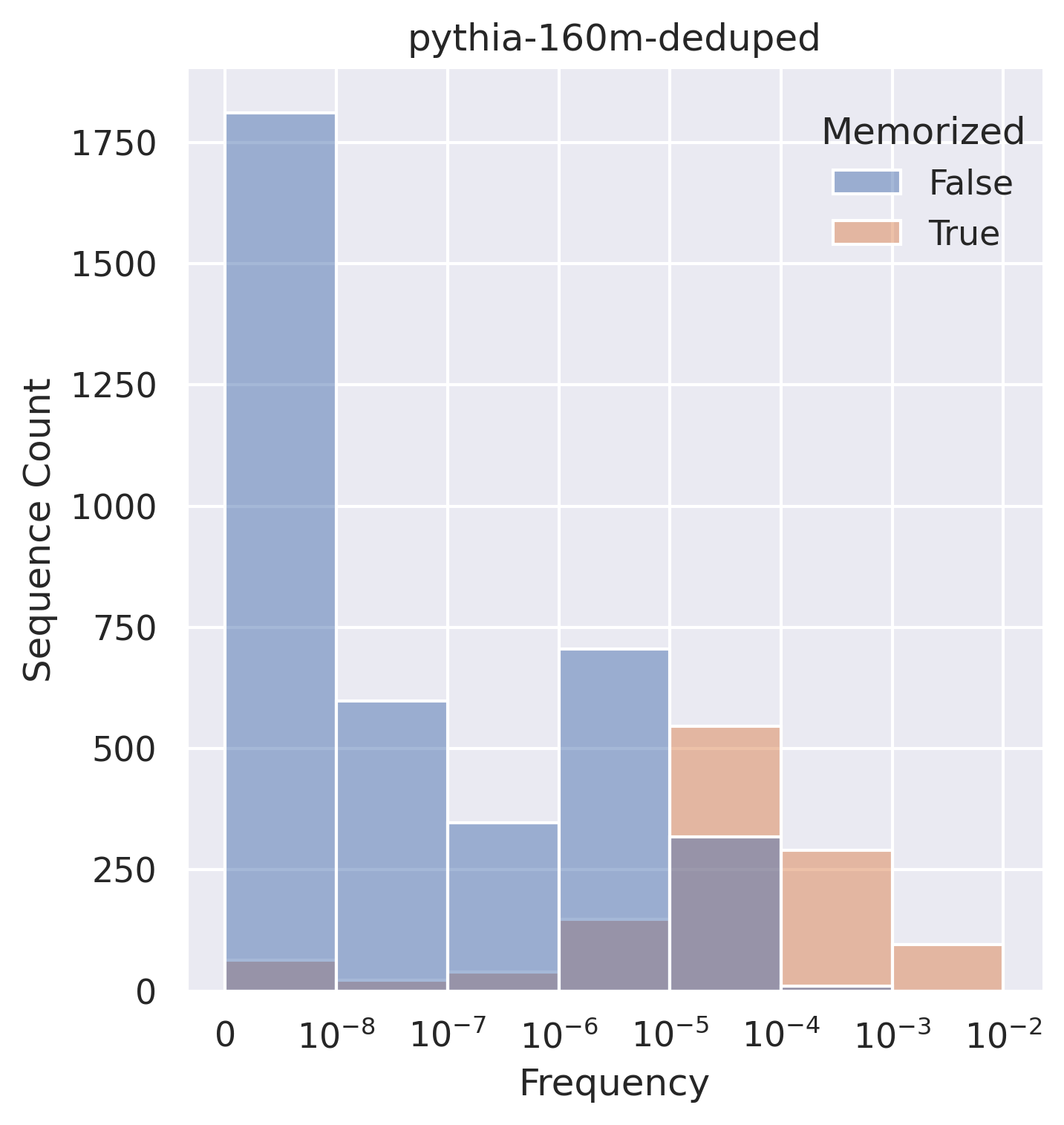}
    \caption{\pythia{160m}}
    \label{fig:exp-pile-freq-160m}
    \end{subfigure}%
    \begin{subfigure}[t]{0.32\linewidth}
\includegraphics[width=\linewidth,trim={5 0 0 20},clip]{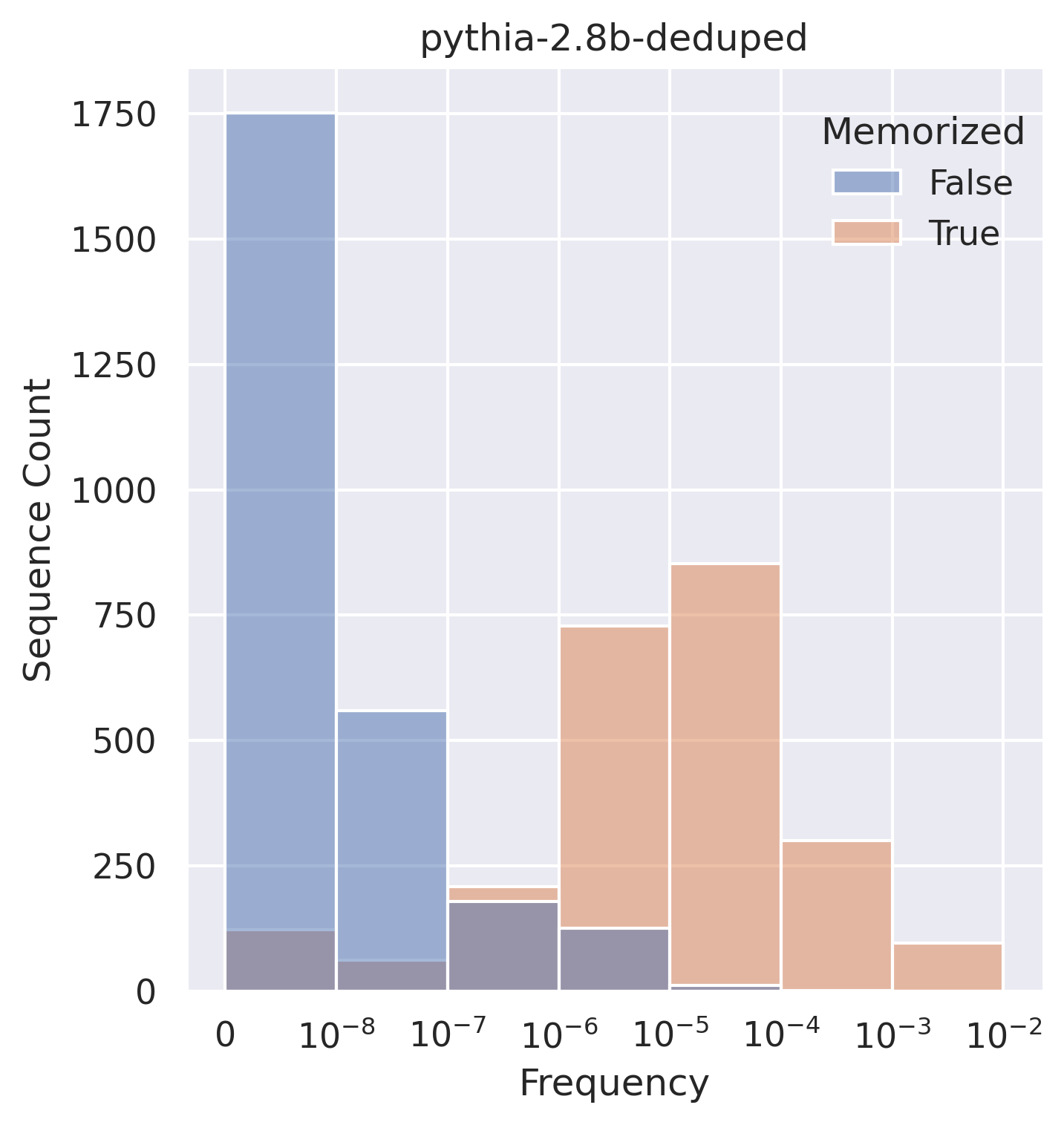}
    \caption{\pythia{2.8b}}
    \label{fig:exp-pile-freq-2.8b}
    \end{subfigure}%
    \begin{subfigure}[t]{0.32\linewidth}
\includegraphics[width=\linewidth,trim={5 0 0 20},clip]{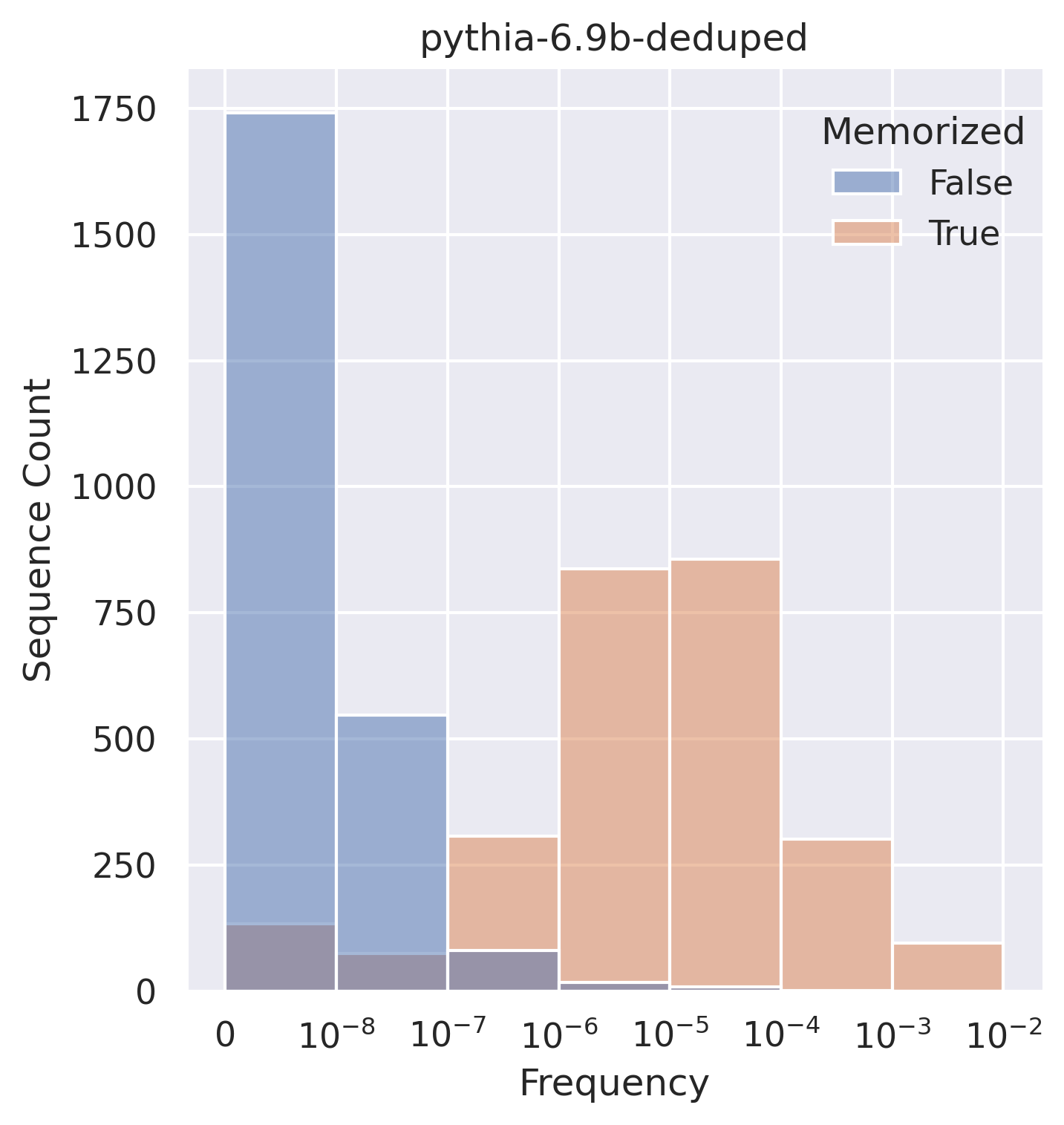}
    \caption{\pythia{6.9b}}
   \label{fig:exp-pile-freq-6.9b}
    \end{subfigure}%
    \caption{Sequence frequency  distribution (on a logarithmic scale) of 2.5K memorized and 2.5K non-memorized sequences, randomly sampled from the 98M Pythia deduped training data. The non-memorized sequences (blue bars) are downsampled 110 times relative to memorized sequences (orange bars).}
    \label{fig:exp-pile-freq}
    \vspace{-3ex}
\end{figure*}

\paragraph{Counting} We uniformly sample about 10M sequences from the Pile deduped dataset to estimate the frequency of each probe. The 10M sequences are sampled at every 10K training step starting at step 0, with a total of 10$\times$1000$\times$1024 sequences of length 2048 tokens.  We count the number of occurrences of each probe in the 10M sequences. These probes indeed cover a wide range of frequencies from 0 to \E{5}{-3}. The full distribution is shown in Figure~\ref{fig:exp-pile-freq}.

\paragraph{Evaluating models} For each model, we measure the verbatim memorization length on the 5K set of probes. The distribution of memorization length is shown in Figure~\ref{fig:exp-pile-freq}. Aligned with findings from~\citet{carlini2023quantifying} and \citet{prashanth2024recitereconstruct}, we observe that, as model size increases, the median frequency of memorized sequences decreases from \E{4}{-5}, \E{1}{-5}, to \E{9}{-6}. As we mainly experiment with the $\texttt{160m}$ model, we choose two frequencies where there is a mix of memorized and non-memorized sequences: \E{2}{-5} (which is at about the bottom 25th percentile, where sequences are more likely to be non-memorized) and \E{1}{-4} (which is around the top 25th percentile, where sequences are very likely memorized). 

\subsection{Examples of the Single-shot Verbatim Memorization Illusion}
\label{appx:single_shot_illusion}

Figure~\ref{fig:example-single-occurrence} shows examples of sequences that only occur in the Pile once or twice according to the infini-gram tool\footnote{\url{https://huggingface.co/spaces/liujch1998/infini-gram}} and would be considered as verbatim memorized based on the most commonly used extractability definition~\cite{carlini2021extracting,carlini2023quantifying,prashanth2024recitereconstruct}, i.e., a memorized sequence of 32 tokens can be extracted with a prefix length of 32 tokens. In reality, these sequences are either under-counted due to limitations of string-based matching or simply not verbatim memorized -- a checkpoint produced before the step where the sequence occurs can already generate a close variant of the sequence.

\begin{figure*}[t!]
    \centering
    
    \fbox{
    \begin{minipage}{0.96\textwidth} 
    \texttt{\textbf{{Template}}} \\
    \texttt{\textcolor{gray}{PROMPT: imGroupExternalMembershipManager getExternal}} \\
    \texttt{\textcolor{gray}{OUTPUT: MembershipManager() \{\textbackslash n        return externalMembershipManager;\textbackslash n    \}\textbackslash n\textbackslash n    public void setExternalMembershipManager}} \\
    \texttt{\textcolor{gray}{PROMPT: madesimple.statoil\textbackslash n\textbackslash ncmsmadesimple.lips}} \\
    \texttt{\textcolor{gray}{OUTPUT: y\textbackslash n\textbackslash ncmsmadesimple.next\textbackslash n\textbackslash ncmsmadesimple.nextdirect\textbackslash n\textbackslash ncmsmadesimple.ubs \textbackslash n\textbackslash ncmsmadesimple.war}} \\
    
    \texttt{\textbf{{Variation}}} \\
    \texttt{\textcolor{gray}{PROMPT: the testimony.\textbackslash n\textbackslash n               Rule 702 which}} \\
    \texttt{\textcolor{gray}{OUTPUT: governs the admissibility of expert testimony provides:\textbackslash n\textbackslash n\textbackslash n           If scientific, technical, or other specialized knowledge will assist the trier of fact\textbackslash n       to understand the}} \\
    \texttt{\textcolor{gray}{PROMPT:  13: Thou shalt tread upon the lion and adder: the young}} \\
    \texttt{\textcolor{gray}{OUTPUT:  lion and the dragon shalt thou trample under feet. 14: Because he hath set his love upon me, therefore will I deliver him: I will set}} \\
    
    \texttt{\textbf{{Induction}}} \\
    \texttt{\textcolor{gray}{PROMPT: 01 3600\textbackslash n265 3586 3587 3602 3601\textbackslash n266 35}} \\
    \texttt{\textcolor{gray}{OUTPUT: 87 3588 3603 3602\textbackslash n267 3588 3589 3604 3603\textbackslash n268 3589 3590 3605 3604\textbackslash n269 3590 3591 3606 3605}} \\
    \texttt{\textcolor{gray}{PROMPT: ang12.bdf batang12b.bdf \textbackslash\textbackslash\textbackslash n\textbackslash t\textbackslash tbatang14.bdf batang14b.bdf batang16.}} \\
    \texttt{\textcolor{gray}{OUTPUT: bdf batang16b.bdf \textbackslash\textbackslash\textbackslash n\textbackslash t\textbackslash tbatang18.bdf batang18b.bdf batang20.bdf batang}} \\
    
    \texttt{\textbf{{Composition}}} \\
    \texttt{\textcolor{gray}{PROMPT: normal; AST: aspartate aminotransferase (i.e. SGOT:}} \\
    \texttt{\textcolor{gray}{OUTPUT:  serum glutamic oxaloacetic transaminase); ALT: alanine aminotransferase (i.e. SGPT: serum glutamic pyruvic transaminase}} \\
    \texttt{\textcolor{gray}{PROMPT: G), tenofovir alafenamide/emtricitabine/bic}} \\
    \texttt{\textcolor{gray}{OUTPUT: tegravir (TAF/FTC/BIC), and tenofovir alafenamide/emtricitabine/ rilpivirine (TAF}}
    \end{minipage}}
    
    \caption{Examples of the single-shot verbatim memorization illusion. Each example is a sequence that occurs once or twice in the \pythia{6.9b} training data and can be generated by the model verbatim. However, these sequences are likely not learned from a single instance or simply not verbatim memorized -- even with a model checkpoint produced before the training step where the memorized sequence occurs, the model can already output the ``memorized'' sequence or a close variant.}
    
    \label{fig:example-single-occurrence}
\end{figure*}

These findings suggest that a model generates a sequence that only occurs once in the training data does not necessarily mean that the model verbatim memorized a sequence after one exposure. As shown in \S\ref{sec:trigger_dependency} and \S\ref{sec:cross_model_intervention}, these sequences may well be ``reconstructed'' by the general language model.

\section{Details of Experiment Setup}
\label{appx:training_setup}

\subsection{Pre-training Data}
We use the Pile deduped version released here,\footnote{\url{https://huggingface.co/datasets/EleutherAI/pile-deduped-pythia-preshuffled}} which contains training data in the exact order they were seen by the Pythia \texttt{deduped} models. For our training runs, we use the data from step $80$K to $82$K, which contain $2$M training examples that has not been seen by any of the checkpoints that we experimented with (except for individual examples with duplicates).

\subsection{Injection Sequences}
\label{appx:inject_data}
\paragraph{Data sources}
We sampled 100 documents from the Internet that are published after Dec 31th 2020, i.e., the Pile corpus cutoff date. These documents are from five sources that have clear publication timestamps: %
Wikipedia,\footnote{\url{https://www.wikipedia.org/}} 
BBC News,\footnote{\url{https://www.bbc.com/news}} 
GitHub,\footnote{\url{https://github.com/}} 
open-access academic papers on ArXiv\footnote{\url{https://arxiv.org/}} and 
Nature,\footnote{\url{https://www.nature.com/}} and quotes from novels.\footnote{\url{https://www.goodreads.com/}} All these sources are covered in the original training corpus. For Wikipedia, we sample articles from 2023 categories curated by Wikipedia, for example, the new product category.\footnote{\url{https://en.wikipedia.org/wiki/Category:Products_introduced_in_2023}} For BBC news, we use the preprocessed corpus on Huggingface.\footnote{\url{https://huggingface.co/datasets/RealTimeData/bbc_news_alltime}} For GitHub, we use code samples from three new programming languages released after 2020: Mojo,\footnote{\url{https://github.com/modularml/mojo}} Gleam,\footnote{\url{https://github.com/gleam-lang/gleam}} and Carbon.\footnote{\url{https://github.com/carbon-language/carbon-lang}}

\paragraph{Verify a sequence is not in the Pile} In our study, an important criterion for injected sequences is that they do not have significant overlap with the pre-training corpus. This is partially ensured by the document publication date. However, we conduct additional verification.

We use two recently open sourced tools that create a searchable index of the Pile. We primarily rely on
Data portraits,\footnote{\url{https://pile.dataportraits.org/}} which directly checks for overlap between a query text and the Pile corpus using Bloom filters computed from 50-character hashes~\cite{marone2023dataportraits}. Bloom filters guarantee no false negatives, however, there will be false positives, i.e., 50-char texts that are not in the Pile but are marked as overlaps. We further confirm these false positives using infini-gram. With both tools, we verify that none of the documents have an overlap with the Pile of more than 50 characters. %

\subsection{Model Checkpoints}
For the sequence injection and the causal dependency experiments, we use the 1K, 10K, 40K, 80K, and the final checkpoints from \pythia{160m}\footnote{\url{https://huggingface.co/EleutherAI/pythia-160m-deduped}}, \pythia{2.8b}\footnote{\url{https://huggingface.co/EleutherAI/pythia-2.8b-deduped}}, and \pythia{6.9b} models.\footnote{\url{https://huggingface.co/EleutherAI/pythia-6.9b-deduped}} For the unlearning stress test evaluation, we follow the setup in~\citet{stoehr2024localizing} and use \texttt{gpt-neo-125m},\footnote{\url{https://huggingface.co/EleutherAI/gpt-neo-125}} which is also pre-trained on the Pile.

\subsection{Setup for the Single-shot Verbatim Memorization Experiment in \S\ref{sec:single_shot_illusion}}
We randomly sample 16 sequences from the 100 injection sequences curated in Appendix~\ref{appx:inject_data}. For each injection sequence, we use the first 224 tokens instead of the full 256 tokens, i.e., a window size of 224, so that we can fit a batch of 32 sequences on a single GPU. In general, with a fixed batch size, a smaller window size makes verbatim memorization more likely to happen, since there are fewer tokens in the batch. Given the actual window size in pre-training is 2048, the verbatim memorization length after a single-shot is likely even smaller than what we observe in our experiment. We experiment with a batch size of 8, 32, and 128.

We use a freshly initialized AdamW optimizer~\cite{loshchilov2018decoupled}, wrapped with the ZeroRedundancyOptimizer,\footnote{\url{https://pytorch.org/tutorials/recipes/zero_redundancy_optimizer.html}} which is used in Pythia training to reduce memory usage~\cite{biderman2023pythia}. The learning rate is set to a constant value of \E{1}{-4}, with other optimizer parameters, e.g., learning rate decay, beta, set to the default values in PyTorch library.\footnote{\url{https://pytorch.org/docs/stable/generated/torch.optim.AdamW.html}} We choose a learning rate that is about twice as large as the actual learning rate at step 80K in pre-training for both the \texttt{2.8b} and \texttt{6.9b} models, with the consideration that a higher learning rate is more likely to produce a large enough gradient update to memorize the injection sequence in a single step.

\subsection{Setup for the Model Quality vs Verbatim Memorization Experiment in \S\ref{sec:checkpoint_vs_mem} and \S\ref{sec:ood}}
\label{appx:training_hyperparam}

Ideally, we want to match the exact pre-training setup that Pythia models used. However, we are constrained by the computation resources available to us. Hence, we choose the closest hyperparameters to the ones used in Pythia pre-training that allow us to fit the model training on a single GPU. Our expectation is that the effects of hyperparameters will be minimized as long as the control and the treatment models use the same set of hyperparameters.

\paragraph{Window size} We use a window size of $256$ tokens for all experiments reported in the paper except the single-shot experiment, a window size that still allows some long range dependencies in the training data. The original window size is $2048$.

\paragraph{Batch size} We use a batch size of 128 examples for the \texttt{160m} models and a batch of size of 40 examples for the \texttt{2.8b} models. These batch size are chosen such that we can fit model training on a single GPU. The original batch size used in pre-training is 1024.

\paragraph{Optimizer state} As we do continued training from different model checkpoints, we experiment with different initial optimizer states and initial learning rates based on the original Pythia learning rate schedule. To initialize the optimizer state, we pre-train from either step 0K (i.e., 1K step before the earliest checkpoint in our experiment) or step 79K (i.e., 1K step before the latest checkpoint in our experiment) for 1M examples. We observe that these two initial states do not affect which checkpoints verbatim memorize more sequences. Thus, when comparing models trained from two different checkpoints, we use the same optimizer state for both. For learning rate, we use the learning rate at the $\texttt{80K}$ checkpoint for each model family, namely \E{2.79}{4} for \texttt{160m} models and \E{7.46}{5} for \texttt{2.8b} models. We observe the learning rate affects all checkpoints equally, with larger learning rates leading to more memorization. We keep the learning rate constant throughout the training, as the amount of data we trained on only corresponds to 1--2K steps in the original training process.

\subsection{Unlearning Method Hyperparameters}
\label{appx:stree_test_setup}
For gradient ascent and sparse fine-tuning, we use the implementation from \citet{stoehr2024localizing}.\footnote{\url{https://github.com/googleinterns/localizing-paragraph-memorization}} We follow the hyperparameters here,\footnote{\url{ https://github.com/googleinterns/localizing-paragraph-memorization/blob/main/notebooks/3\%20editing/fine-tuning.ipynb}} namely, we run optimization for 10 steps using a learning rate of \E{1}{-5} and a weight decay of 1.0. For sparse fine-tuning, we only fine-tune 0.1\% of weights with the highest gradient.

For neuron pruning, we use the implementation from \citet{chang2024localization}.\footnote{\url{https://github.com/terarachang/MemPi}} We prune 0.1\% of the neurons. The L1 penalty is set to 1000. We find that higher L1 penalty leads to degeneration. We run optimization for 1000 steps using a learning rate of \E{1}{-2}. This set of hyperparameters leads to a $\Delta$ self-accuracy of $-$0.248 and $\Delta$ neg-accuracy of $-$0.094 on the 90 sequences to unlearn.

\subsection{Computation Cost} All models are trained on NVIDIA A100 GPUs. For models in \S\ref{sec:single_shot_illusion}, the training is distributed across multiple GPUs, with a local batch size of 32. For models in \S\ref{sec:checkpoint_vs_mem} and \S\ref{sec:ood}, the training is on a single GPU. The training of  \texttt{2.8b} models over 1M examples takes about 16 hours, while the training of  \texttt{160m} models takes about 3 hours.

\section{Additional Experiment Results}

\subsection{Additional Results on Checkpoint vs. Verbatim Memorization Length}
\label{appx:ckpt_vs_mem}

\begin{figure}[ht]
    \centering
    \includegraphics[width=0.7\linewidth,trim={0 0 0 0},clip]{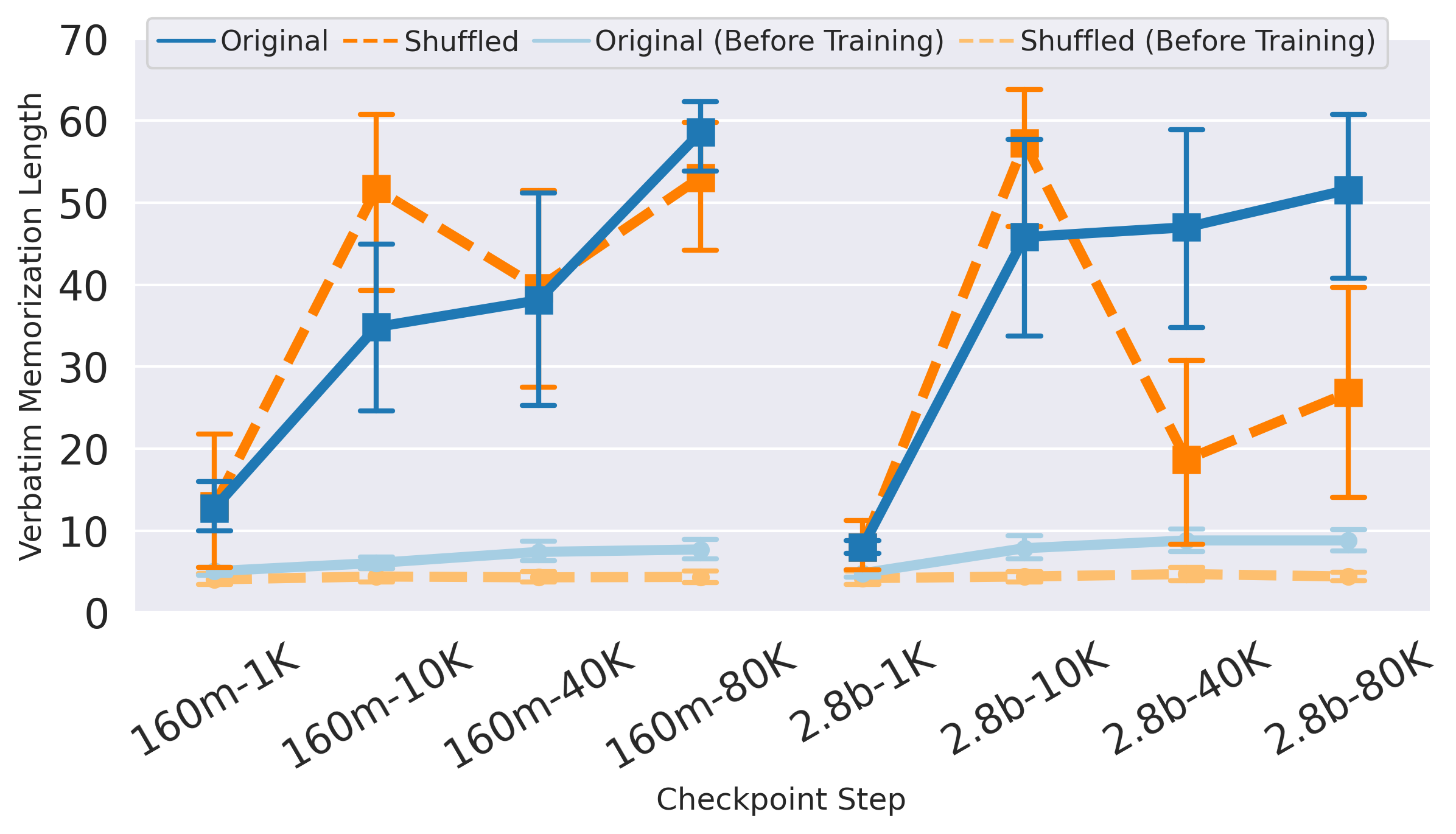}
    \caption{Checkpoint vs.~verbatim memorization length of original and shuffled sequences, with a sequence frequency of every 10K examples.}
    \label{fig:ckpt-vs-mem-1e4}
\end{figure}

In Figure~\ref{fig:ckpt-vs-mem-1e4}, we show the results of verbatim memorization length when continue pre-training from different checkpoints of the \texttt{160m} model and the \texttt{2.8b} model with a sequence injection frequency of 1~in~10K examples. This is a frequency that both models are expected to memorize most of the injection sequences. We still see the consistent trend that we observed in \S\ref{sec:checkpoint_vs_mem} on in-domain sequences: later checkpoints memorize longer sequences. The gap between shuffled sequences and original sequences is narrowed, especially on the \texttt{160m} models, possibly because the model is seeing the injection sequences more frequently. For the \texttt{2.8b} models, which see the injection sequences fewer times, shuffled sequences are still harder to memorize than the original ones for all checkpoints except the 10K step.

\subsection{Additional Results on Causal Dependencies}
\label{appx:dependency}

\paragraph{Behavioral evidence of verbatim memorization is triggered by abstract model states}

\begin{figure*}[t!]
    \centering
    \fbox{
    \begin{minipage}{0.96\textwidth} 
    \texttt{\textbf{{The Original Trigger Prefix}}} \\
    \texttt{\textcolor{gray}{Mr and Mrs Dursley, of}  \textcolor{forestgreen}{number four, Privet Drive, were proud to say that they were perfectly normal, thank you very much}} \\

    \texttt{\textbf{{Trigger Prefixes with Similar High-level Features}}} \\
\texttt{\textcolor{gray}{Mrs and Mr Dursley, of number four, Privet Drive, were proud to say that they were} \textcolor{forestgreen}{perfectly normal, thank you very much}} \\
\texttt{\textcolor{gray}{The Dursley family, of number four, Privet Drive, were proud to say that they were} \textcolor{forestgreen}{perfectly normal, thank you very much}} \\
\texttt{\textcolor{gray}{Mr and Mrs Weasley, residing at four Privet Drive, were proud to say they were} \textcolor{forestgreen}{ perfectly normal, thank you very much}} \\
\texttt{\textcolor{gray}{Mr and Mrs Slytherin, of number twenty-one, Privet Drive, were proud to say that they were} \textcolor{forestgreen}{perfectly normal, thank you very much}} \\
\texttt{\textcolor{gray}{Mr and Mrs Dursley, of \#4, Privet Drive, were proud to say that they were} \textcolor{forestgreen}{ perfectly normal, thank you very much}} \\
\texttt{\textcolor{gray}{Mr and Mrs Dursley, of number ten, Privet Drive, were proud to say that they were} \textcolor{forestgreen}{perfectly normal, thank you very much}} \\
\texttt{\textcolor{gray}{Mr and Mrs Dursley, of Privet Drive, were proud to say that they were} \textcolor{forestgreen}{ perfectly normal, thank you very much}} \\
\texttt{\textcolor{gray}{Mr and Mrs Dursley, of number four, Oak Street, were proud to say that they were} \textcolor{forestgreen}{perfectly normal, thank you very much}} \\
\texttt{\textcolor{gray}{Mr and Mrs Dursley, residing at four Privet Drive, were delighted to assert they were} \textcolor{forestgreen}{ perfectly normal, thank you very much}} \\
\texttt{\textcolor{gray}{The Dursley family, of number four, Privet Drive, were pleased to declare that they were} \textcolor{forestgreen}{ perfectly normal, thank you very much}} \\

    \texttt{\textbf{{Non-Trigger Prefixes with Similar or Different High-level Features}}} \\
\texttt{\textcolor{gray}{Mr and Mrs Kingsley, of number four, Privet Drive, were proud to say that they were} \textcolor{brickred}{ the proud parents of a bouncing baby boy.
}} \\
\texttt{\textcolor{gray}{Mr and Mrs Weasley, of number four, Privet Drive, were proud to say that they were} \textcolor{brickred}{ expecting their first child.}} \\
\texttt{\textcolor{gray}{Mr and Mrs Dursley, of number four, Privet Drive, were glad to say that they were} \textcolor{brickred}{ only too delighted to have the young man staying with them.}} 
    \end{minipage}}
    
    \caption{Examples of trigger prefixes that lead to similar abstract states, i.e., similar high-level semantic features. The \texttt{\textcolor{gray}{gray texts}} are the prompts. The \texttt{\textcolor{forestgreen}{green texts}} are the memorized continuations.  The \texttt{\textcolor{brickred}{red texts}} are the non-memorized continuations.}
    
    \label{fig:example-causal-dependencies-abstract-states}
    \vspace{-2ex}
\end{figure*}

\begin{figure}[ht!]
    \centering
   \includegraphics[width=0.41\linewidth,angle=0]{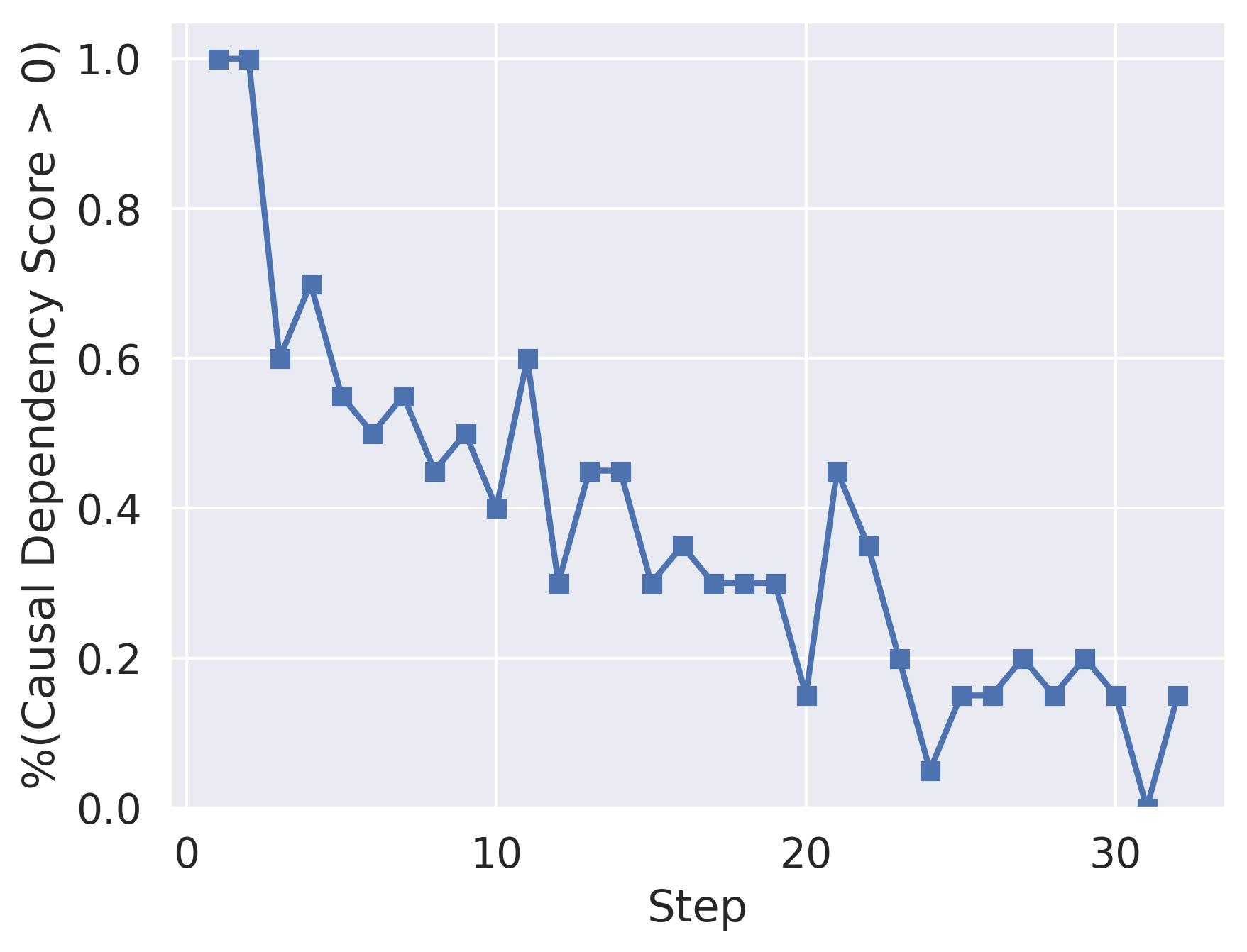}
\includegraphics[width=0.4\linewidth,angle=0]{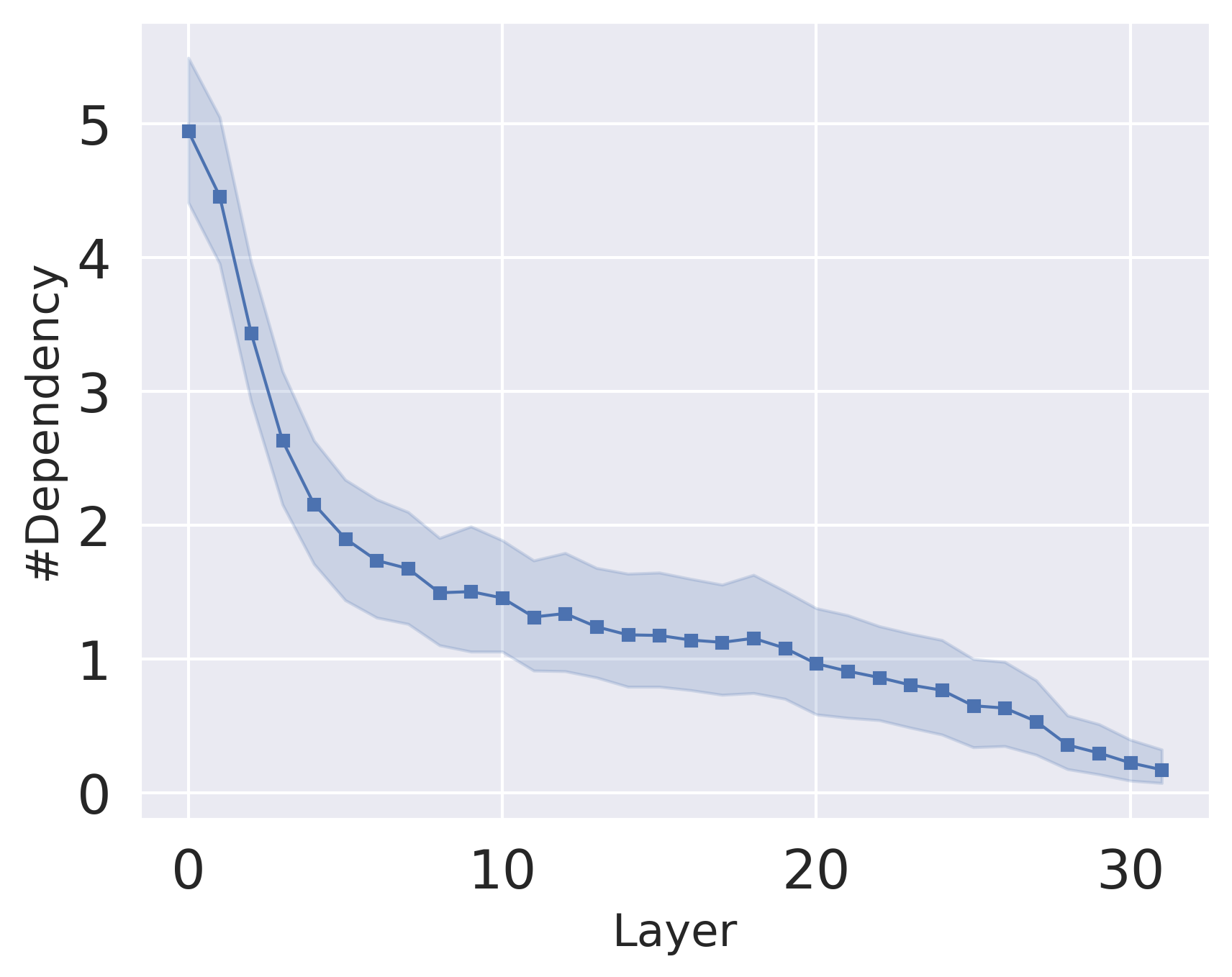}
     \caption{Causal dependencies between memorized tokens and tokens in the trigger for the \texttt{6.9b} model.}
    \label{fig:exp-dep-6.9b}
\end{figure}

In Figure~\ref{fig:example-causal-dependencies-abstract-states}, we show that when prompted with prefixes sharing similar high-level features,  e.g., synonyms or proper nouns belong to the same category, the \texttt{6.9b} model can produce the memorized continuation. Semantically similar prefixes do not always trigger verbatim memorization, nor does verbatim memorization strictly require prefixes semantically similar to the one in training, however, semantically relevant substitutions do have a higher probability to trigger the verbatim memorized continuation than random substitutions.

Overall, the trigger is a set of \emph{distributed abstract states} and does not require a particular token to be presented in the prefix, i.e., the verbatim memorization is not triggered by a single n-gram match. This finding motivates the stress tests we developed in \S\ref{sec:stress_testing}.

\paragraph{Results of the \texttt{6.9b} model}
In Figure~\ref{fig:exp-dep-6.9b}, we show the causal dependency results of the pre-trained \pythia{6.9b} model on 50 memorized sequences sampled from the Pile. The results are consistent with what we observed from the \texttt{160m} models trained using our sequence injection framework -- namely, not all verbatim memorized tokens depend on the trigger sequences. Moreover, for memorized tokens that depend on the trigger, the dependencies are also around middle layers, suggesting high-level features are involved in the verbatim memorization.

\subsection{Verbatim Memorization Can Still Happen with Frozen Attention Heads}
\label{appx:frozen_components}

\begin{figure}[t]
    \centering
   \includegraphics[width=0.45\linewidth,angle=0]{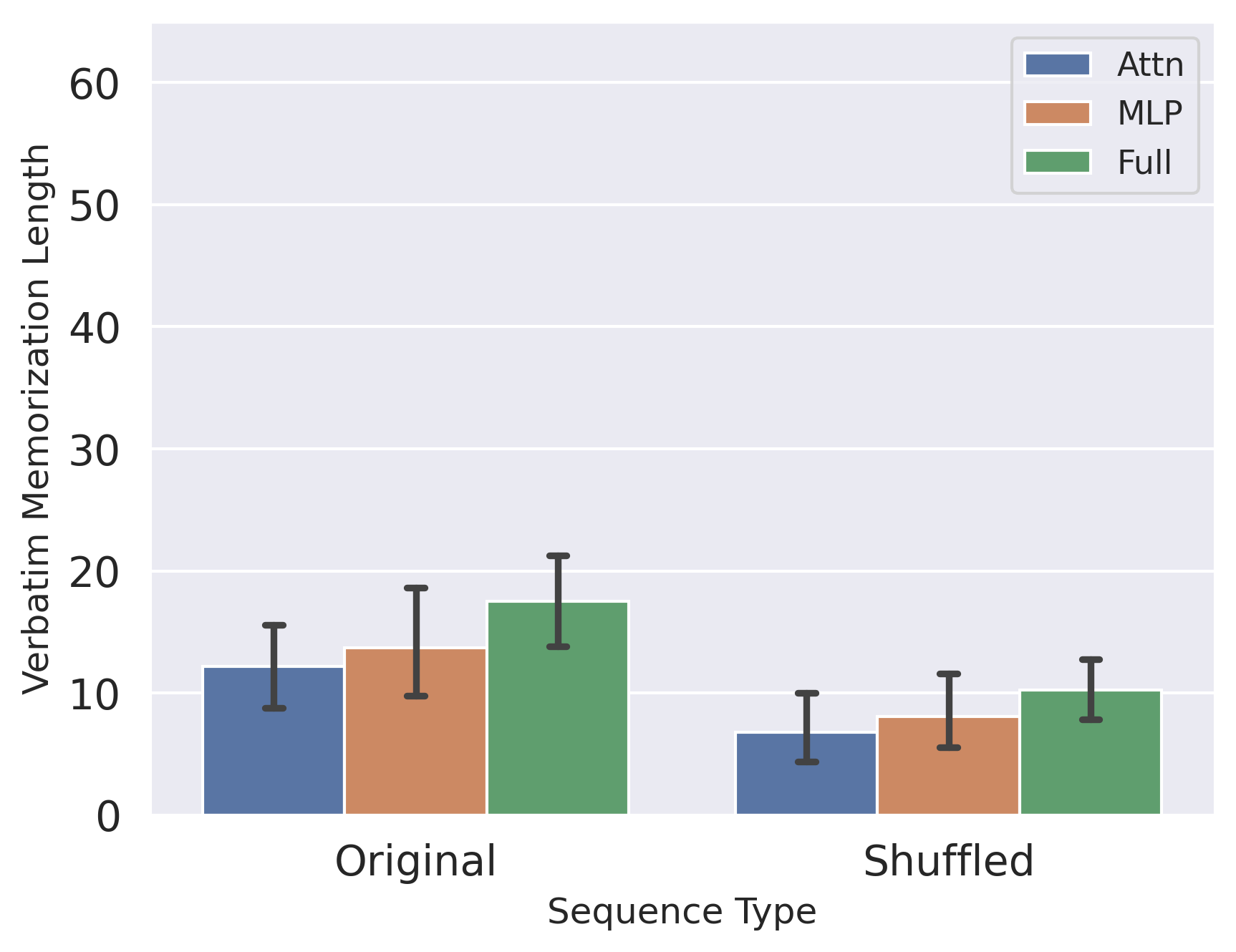}
   \includegraphics[width=0.45\linewidth,angle=0]{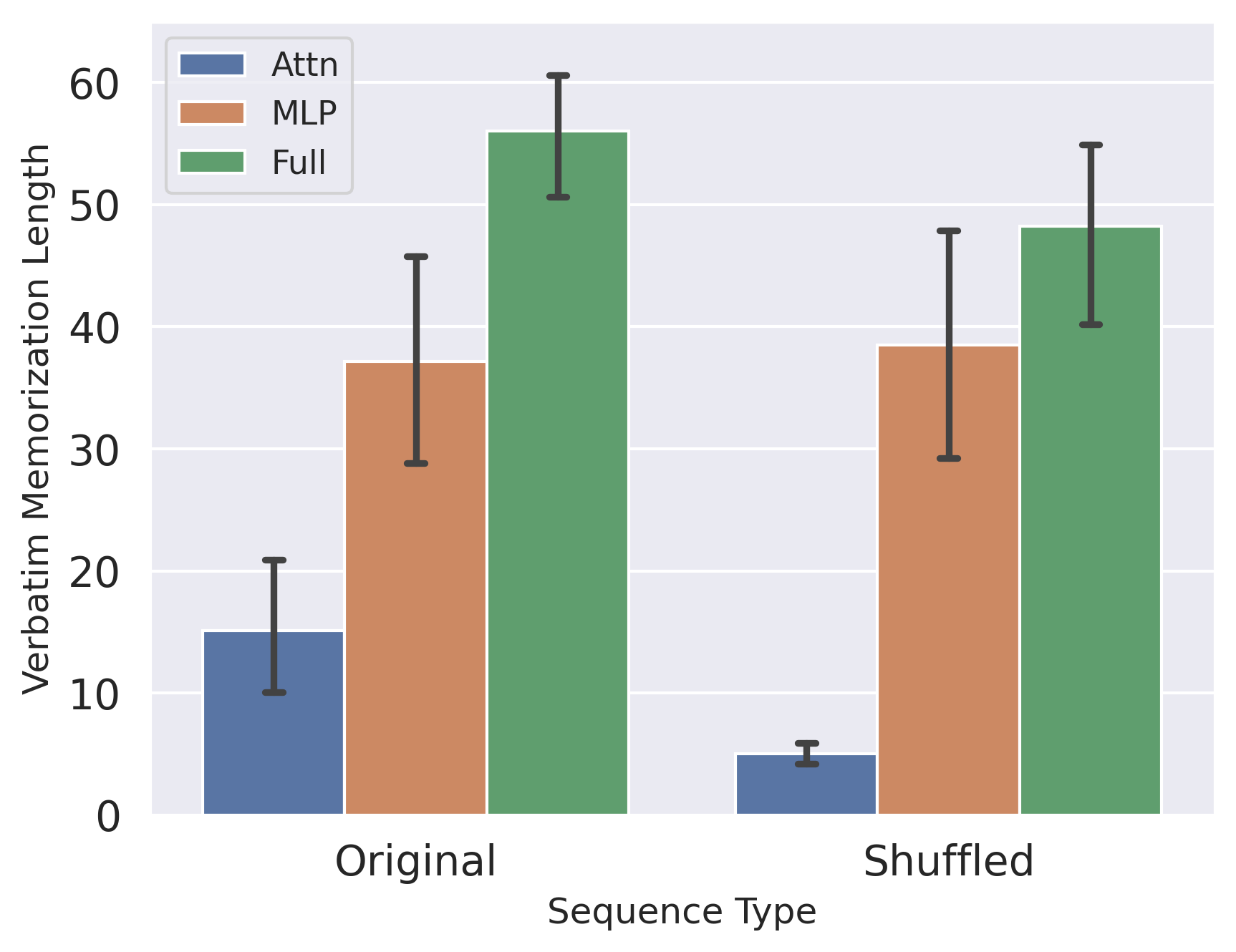}
     \caption{Trainable components vs.~verbatim memorization length of original and shuffled sequences. Models are trained from the \texttt{160m} model checkpoint at step 80K with two different data injection frequencies (every 50K and 10K examples).}
    \label{fig:exp-component-vs-mem}
\end{figure}

Experiments in \S\ref{sec:cross_model_intervention} show that both attention and MLP components are involved in verbatim memorization. We now investigate which components are strictly necessary for verbatim memorization, taking the capacity of these components into account.

\paragraph{Setup} We conduct an ablation study by training three sets of models using the \texttt{160m} model checkpoints: (1) only MLP layer weights are trainable (2) only attention head weights are trainable (3) all weights are trainable. For Pythia models, which use the GPTNeoX architecture, the MLP layers contain 35\% of model weights while attention heads contain 17\% of model weights. We experiment with two sequence frequency of 1~in~50K and 1~in~10K.

\paragraph{Results}
Results are shown in Figure~\ref{fig:exp-component-vs-mem}. With a frequency of every 50K examples, neither MLP-only nor attention-only models can verbatim memorize a sequence. However, at the frequency of every 10K examples, the model with frozen attention heads surprisingly can verbatim memorize sequences of 40 tokens on average, which is about 80\% of the tokens memorized by a model with all weights trainable. These results suggest that MLP layers are strictly necessary for verbatim memorization, while attention mechanisms learned from general language modeling can largely be reused assuming the sequence to memorize occurs frequently enough in training.

\subsection{Additional Results on Stress Testing Unlearning Methods}
\label{appx:unlearning}

\begin{figure*}[!t]
    \small
    \centering
    \begin{subfigure}[t]{0.25\linewidth}
\includegraphics[width=\linewidth,trim={0 0 0 0},clip]{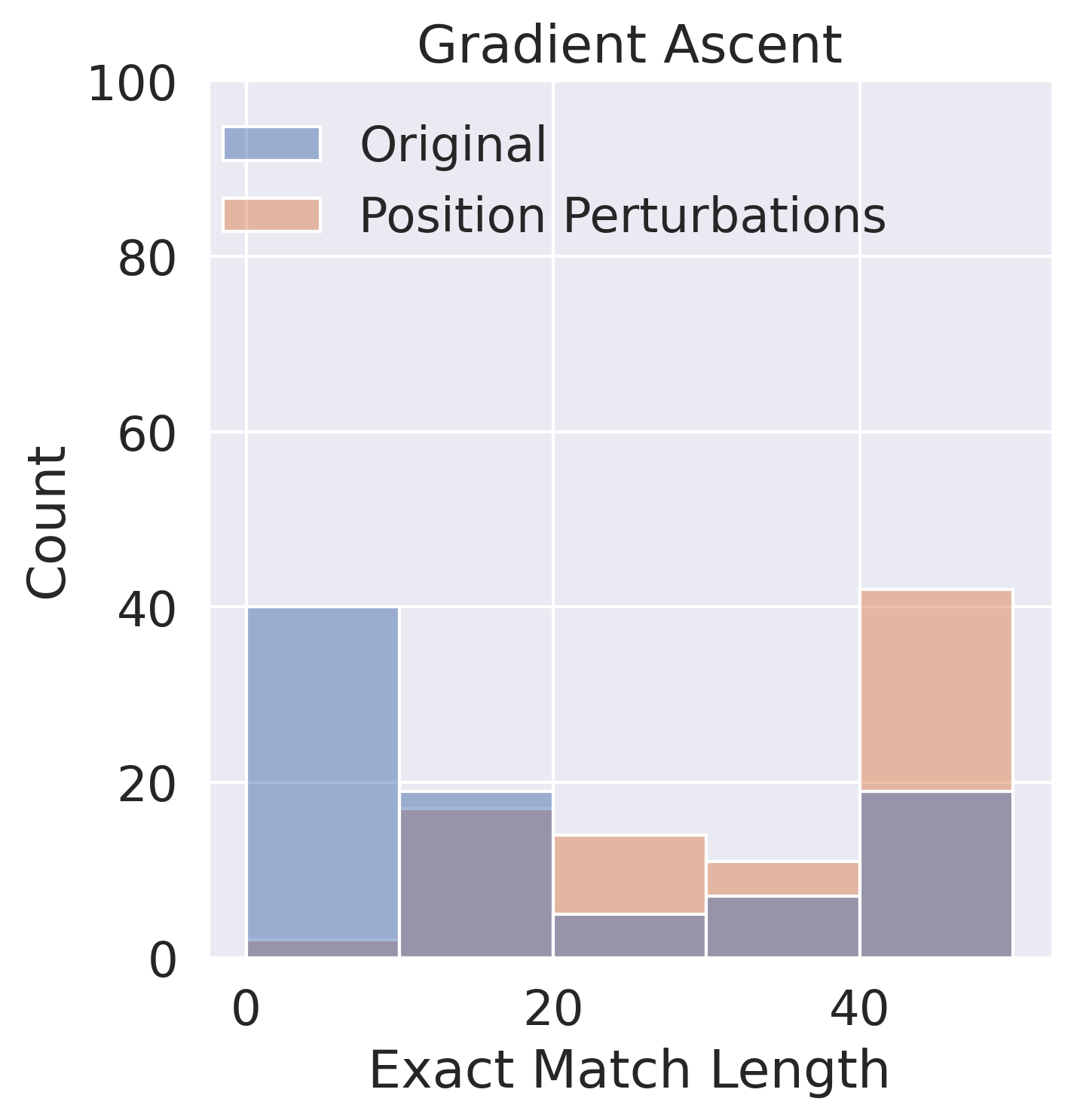}
    \end{subfigure}%
    \begin{subfigure}[t]{0.25\linewidth}
\includegraphics[width=\linewidth,trim={0 0 0 0},clip]{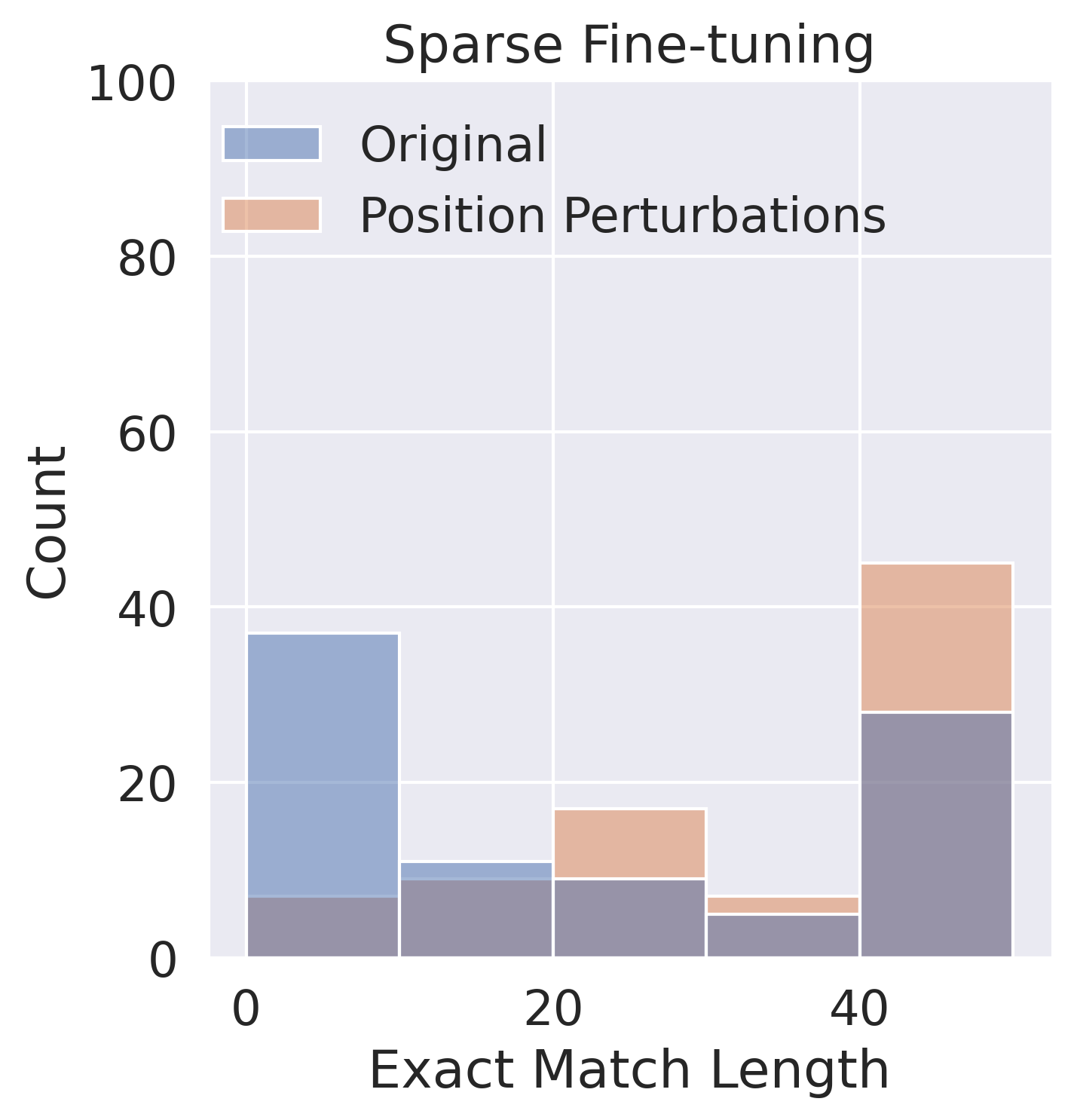}
    \end{subfigure}%
    \begin{subfigure}[t]{0.25\linewidth}
\includegraphics[width=\linewidth,trim={0 0 0 0},clip]{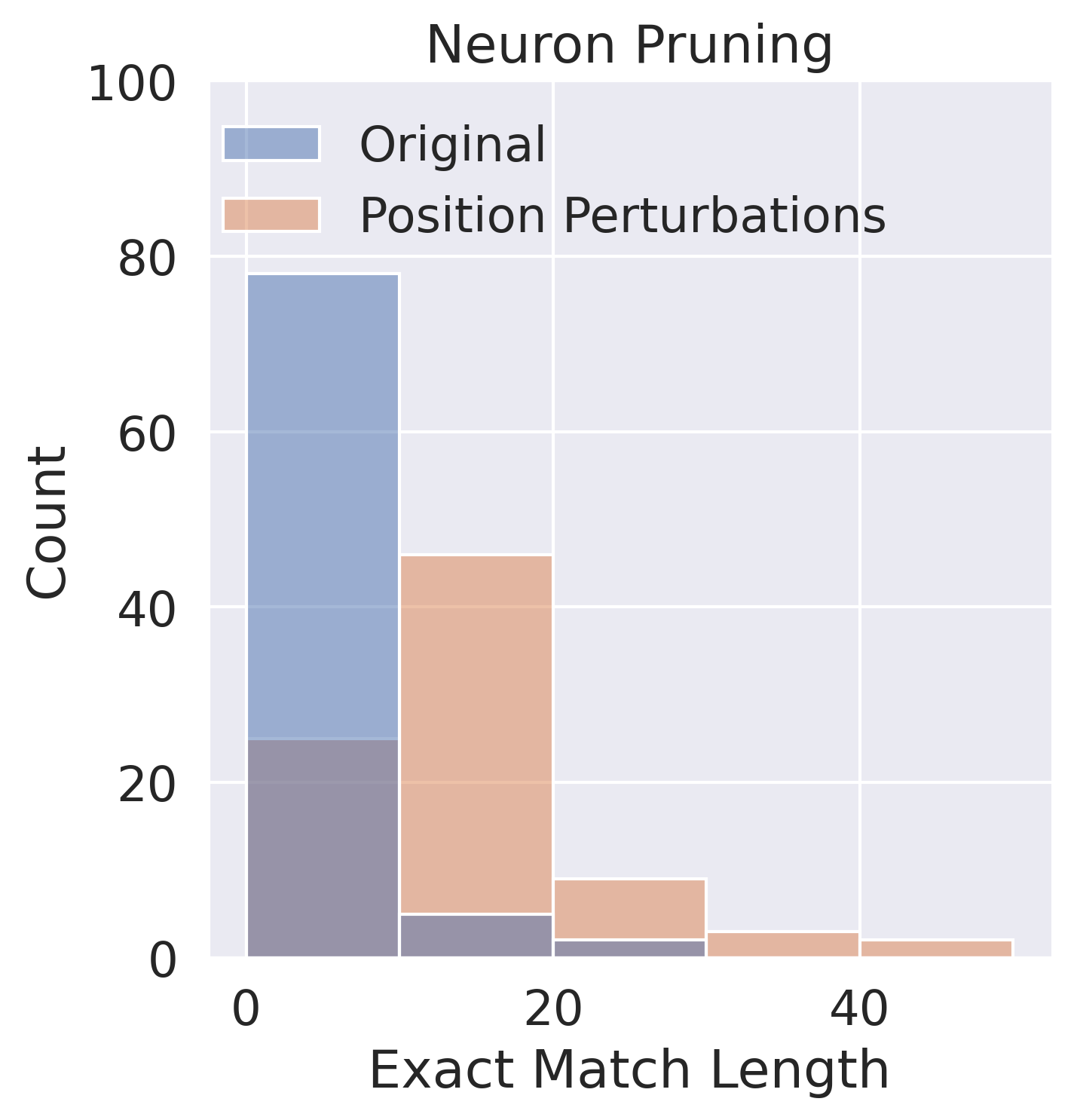}
    \end{subfigure}%
    \\
    \begin{subfigure}[t]{0.25\linewidth}
\includegraphics[width=\linewidth,trim={0 0 0 0},clip]{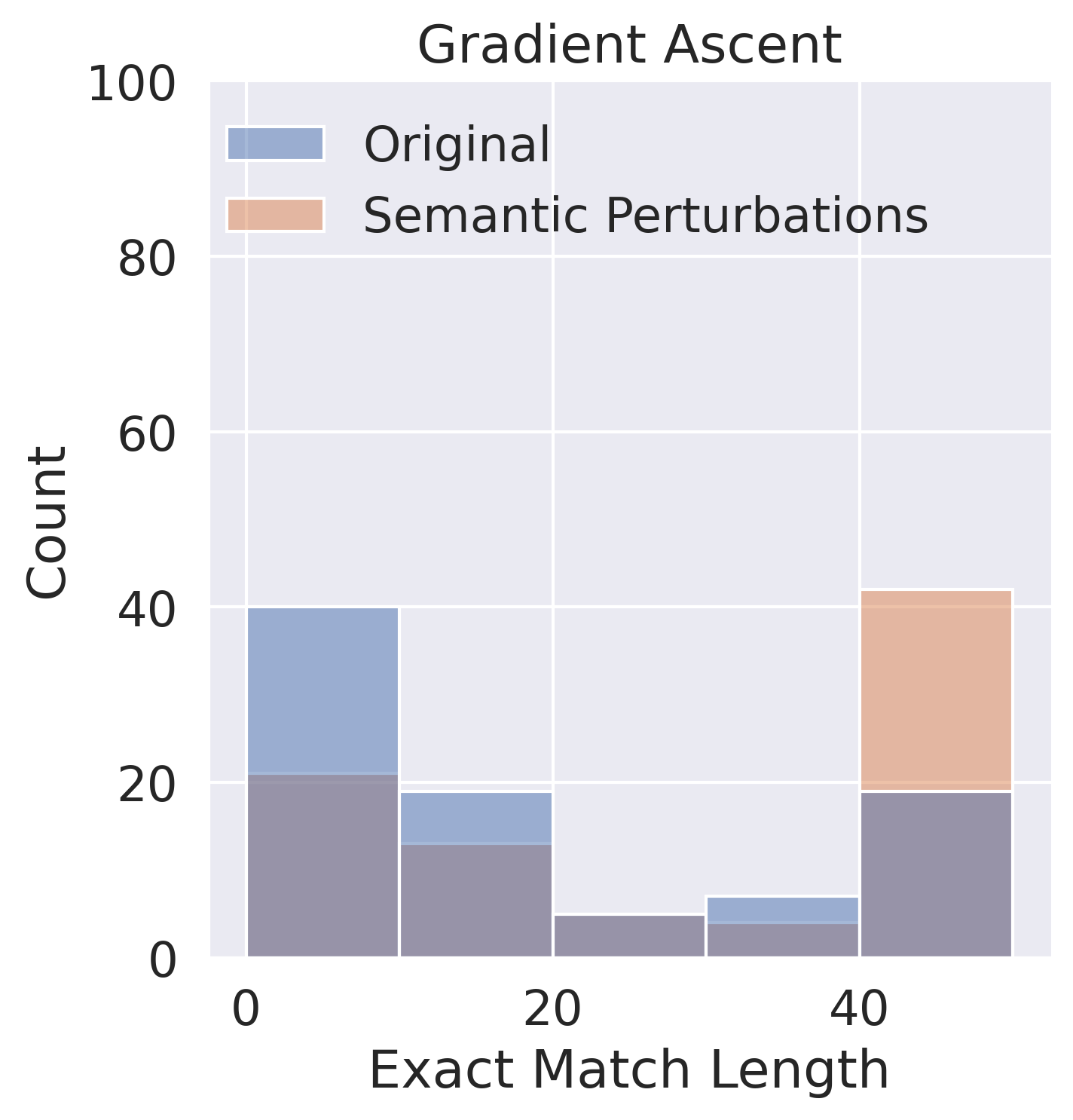}
    \end{subfigure}%
    \begin{subfigure}[t]{0.25\linewidth}
\includegraphics[width=\linewidth,trim={0 0 0 0},clip]{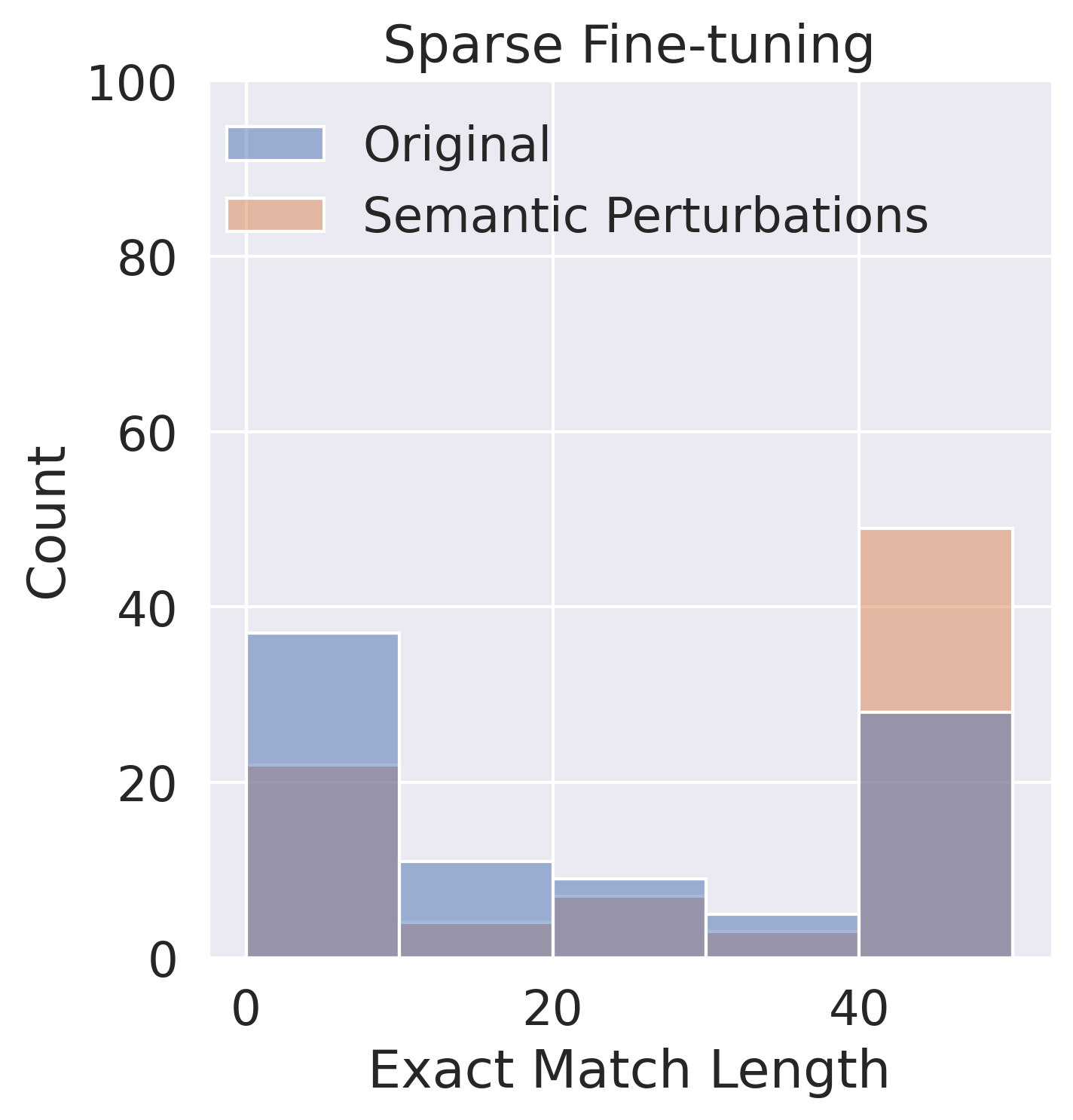}
    \end{subfigure}%
    \begin{subfigure}[t]{0.25\linewidth}
\includegraphics[width=\linewidth,trim={0 0 0 0},clip]{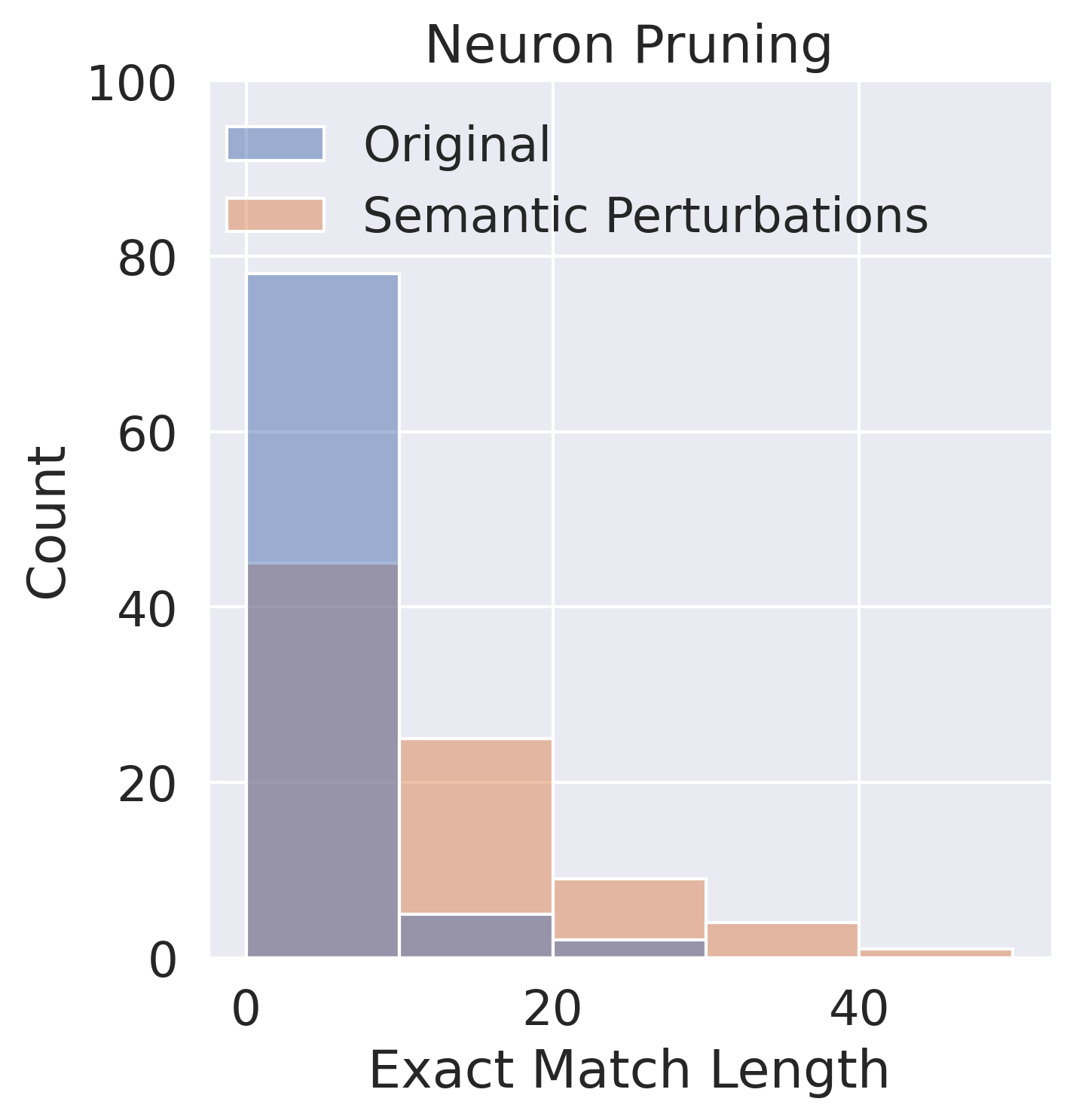}
    \end{subfigure}%
    \caption{Verbatim memorization length distribution on the original prompts and the stress testing prompts.}
    \label{fig:exp-unlearning-dist}
    \vspace{-3ex}
\end{figure*}

In Figure~\ref{fig:example-stress-test}, we show examples of the original prompt and the perturbed stress testing prompts, along with the model output before and after unlearning. In Figure~\ref{fig:exp-unlearning-dist}, we show the verbatim memorization length distribution shift when evaluate with the original prompts and the stress testing prompts. 

\section{License}

For artifacts used in this work, the Pythia models are licensed under Apache-2.0 License. The Pile dataset is licensed under MIT License. Our use of the model and the dataset are permitted under the license.

\begin{figure}[t!]
    \centering
    
    \fbox{
    \begin{minipage}{0.96\linewidth} 

    \texttt{Unlearning with Gradient Ascent} \\
    \texttt{\textbf{{Original Test}}} \\
    \texttt{\textcolor{gray}{PROMPT: 0f86e5b48e01b996cadc001622fb5e363b421",\textbackslash n            "uncleHash" : "0x1dcc4de8}} \\
    \texttt{\textcolor{gray}{OUTPUT: dec75d7aab85b567b6ccd41ad312451b948a7413f0a142fd40d49347"\textbackslash n        \},\textbackslash n      }} \\
    \texttt{\textcolor{brass}{UNLEARNED: 67\textbf{b567b6ccd41ad312451b9}24af0a0af82a49f0d2c4934\textbackslash n        \}\textbackslash n    \}\textbackslash n\}\textbackslash n2:\{}} \\
    \texttt{\textbf{{Position Perturbations}}} \\
    \texttt{\textcolor{gray}{PROMPT: 5b48e01b996cadc001622fb5e363b421",\textbackslash n            "uncleHash" : "0x1dcc4de8}} \\
    \texttt{\textcolor{gray}{OUTPUT: dec75d7aab85b567b6ccd41ad312451b948a7413f0a142fd40d49347"\textbackslash n        \},\textbackslash n      }} \\
    \texttt{\textcolor{brickred}{UNLEARNED: \textbf{dec75d7aab85b567b6ccd41ad312451b948a7413f0a142fd40d49347"\textbackslash n        \},\textbackslash n      }}} \\
    \texttt{\textbf{{Semantic Perturbations}}} \\
    \texttt{\textcolor{gray}{PROMPT: e105b48e01b996cadc001622fb105e363b421",\textbackslash n            "uncleHash" : "0x1dcc4de8}} \\
    \texttt{\textcolor{gray}{OUTPUT: dec75d7aab85b567b6ccd41ad312451b948a7413f0a142fd40d49347"\textbackslash n        \},\textbackslash n      }} \\
\texttt{\textcolor{brickred}{UNLEARNED: \textbf{dec75d7aab85b567b6ccd41ad312451b948a7413f0a142fd40d49347"\textbackslash n        \},\textbackslash n      }}}
    
    \noindent\makebox[\textwidth]{\rule{\textwidth}{0.4pt}}

    \texttt{Unlearning with Gradient Ascent} \\
    \texttt{\textbf{{Original Test}}} \\
    \texttt{\textcolor{gray}{PROMPT: NOT TO BE PUBLISHED IN OFFICIAL REPORTS\textbackslash n California Rules of Court, rule 8.1115(a), prohibits courts and parties from citing or relying on opinions not certified for\textbackslash n publication}} \\
    \texttt{\textcolor{gray}{OUTPUT: or ordered published, except as specified by rule 8.1115(b). This opinion has not been certified for publication\textbackslash n or ordered published for purposes of rule 8.1115.\textbackslash n\textbackslash n\textbackslash n}} \\
    \texttt{\textcolor{brass}{UNLEARNED: \textbf{or ordered published, except as specified by rule 8.} coli. This Court has not. All opinions are not treated as a whole, and hence opinions are not certified for purposes of\textbackslash n publication.\textbackslash n}} \\
    \texttt{\textbf{{Position Perturbations}}} \\
    \texttt{\textcolor{gray}{PROMPT: NOT TO BE PUBLISHED IN OFFICIAL REPORTS\textbackslash n California Rules of Court, rule 8.1115(a), prohibits courts and parties from citing or relying on opinions not certified for\textbackslash n publication or ordered published, except as specified by rule 8.11}} \\
    \texttt{\textcolor{gray}{OUTPUT: 15(b). This opinion has not been certified for publication\textbackslash n or ordered published for purposes of rule 8.1115.\textbackslash n\textbackslash n\textbackslash n}} \\
    \texttt{\textcolor{brickred}{UNLEARNED: \textbf{15(b). This opinion has not been certified for publication\textbackslash n or ordered published for purposes of rule 8.1115.\textbackslash n\textbackslash n\textbackslash n}}} \\
    \texttt{\textbf{{Semantic Perturbations}}} \\
    \texttt{\textcolor{gray}{PROMPT: stay PUBLISHED IN OFFICIAL REPORTS\textbackslash n California Rules of Court, rule 8.1115(a), prohibits courts and parties from citing or relying on opinions not certified for\textbackslash n publication}} \\
    \texttt{\textcolor{gray}{OUTPUT: or ordered published, except as specified by rule 8.1115(b). This opinion has not been certified for publication\textbackslash n or ordered published for purposes of rule 8.1115.\textbackslash n\textbackslash n\textbackslash n}} \\
    \texttt{\textcolor{brickred}{UNLEARNED: \textbf{or ordered published, except as specified by rule 8.1115(b). This opinion has not been certified for publication\textbackslash n or ordered published for purposes of rule 8.1115.\textbackslash n\textbackslash n\textbackslash n}}} \\
    \end{minipage}}

    \caption{Examples of stress tests and failure cases of unlearning. Each example consists of the original prompts and the perturbed prompts, along with the model outputs before and after unlearning. Overlaps between the memorized output and the unlearned output are bolded. These examples show that while unlearning largely prevents the model from outputting the memorized sequence given the original prompt, it does not fully remove the verbatim memorized information.}
    
    \label{fig:example-stress-test}
\end{figure}

\begin{figure}[t!]\ContinuedFloat
    \centering
    
    \fbox{
    \begin{minipage}{0.96\linewidth} 

    \texttt{Unlearning with Neuron Pruning} \\
    \texttt{\textbf{{Original Test}}} \\
    \texttt{\textcolor{gray}{PROMPT: From fairest creatures we desire increase,\textbackslash n That thereby beauty’s rose might never die.\textbackslash n But as the riper should by time decease}} \\
    \texttt{\textcolor{gray}{OUTPUT: ,\textbackslash n His tender heir might bear his memory:\textbackslash n But thou, contracted to thine own bright eyes,\textbackslash n Feed’st thy light's flame with self-substantial fuel,\textbackslash n Making a famine where abundance lies,\textbackslash n Thyself thy foe}} \\
    \texttt{\textcolor{forestgreen}{UNLEARNED: ,\textbackslash n So too our own, to our own selves, shall grow\textbackslash n\textbackslash n The world is a book, and those who do not travel read only a page.\textbackslash n\textbackslash n The world is a book, and those who do not travel read only}} \\
   \texttt{\textbf{{Position Perturbations}}} \\
    \texttt{\textcolor{gray}{PROMPT: From fairest creatures we desire increase,\textbackslash n That thereby beauty’s rose might never die.\textbackslash n But as the riper should by time decease,\textbackslash n His tender heir might bear his memory:\textbackslash n But thou, contracted to thine own bright eyes,\textbackslash n Feed’st thy light's}} \\
    \texttt{\textcolor{gray}{OUTPUT:  flame with self-substantial fuel,\textbackslash n Making a famine where abundance lies,\textbackslash n Thyself thy foe}} \\
    \texttt{\textcolor{brickred}{UNLEARNED: \textbf{ flame with self-substantial fuel,\textbackslash n Making a famine where abundance lies,\textbackslash n Thyself thy foe}}} \\
    \texttt{\textbf{{Semantic Perturbations}}} \\
    \texttt{\textcolor{gray}{PROMPT: From fairest creatures we desire increase,\textbackslash n That thereThrough beauty’s rose might never die.\textbackslash n But as the riper should by time decease}} \\
    \texttt{\textcolor{gray}{OUTPUT: \textbackslash n His tender heir might bear his memory:\textbackslash n But thou, contracted to thine own bright eyes,\textbackslash n Feed’st thy light’ning fires with self-substantial fuel;\textbackslash n Making a famine where abundance lies,\textbackslash n Thyself thy foe}} \\
\texttt{\textcolor{brass}{UNLEARNED: \textbf{,\textbackslash n His tender heir might bear his memory:\textbackslash n But} you, whose beauty is for aye the same,\textbackslash n O, you must not dearer be than you are!\textbackslash n\textbackslash n The poem is a parody of the famous "Ode}}

    \end{minipage}}
    
    \caption{Examples of stress tests and failure cases of unlearning methods (cont.).}
    
\end{figure}

\label{sec:appendix}

\end{document}